\documentclass[11pt]{article}
\usepackage{fullpage}
\usepackage{url}
\usepackage{smile}
\usepackage{pdfpages}
\usepackage{kpfonts}

\usepackage[colorlinks=true,
 linkcolor=blue,
 urlcolor=blue,
 citecolor=blue,
 plainpages=false]{hyperref}

\usepackage{tgpagella}
\DeclareMathAlphabet{\mathsf}{OT1}{cmss}{m}{n}
\SetMathAlphabet{\mathsf}{bold}{OT1}{cmss}{bx}{n}

\makeatletter

\newcommand{\Rmnum}[1]{\expandafter\@slowromancap\romannumeral #1@}
\makeatother
\usepackage{natbib}
\numberwithin{equation}{section}
\numberwithin{theorem}{section}
\usepackage[left]{lineno}
\usepackage{blindtext}

\newcommand*\patchAmsMathEnvironmentForLineno[1]{%
  \expandafter\let\csname old#1\expandafter\endcsname\csname #1\endcsname
  \expandafter\let\csname oldend#1\expandafter\endcsname\csname end#1\endcsname
  \renewenvironment{#1}%
     {\linenomath\csname old#1\endcsname}%
     {\csname oldend#1\endcsname\endlinenomath}}% 
\newcommand*\patchBothAmsMathEnvironmentsForLineno[1]{%
  \patchAmsMathEnvironmentForLineno{#1}%
  \patchAmsMathEnvironmentForLineno{#1*}}%
\AtBeginDocument{%
\patchBothAmsMathEnvironmentsForLineno{equation}%
\patchBothAmsMathEnvironmentsForLineno{align}%
\patchBothAmsMathEnvironmentsForLineno{flalign}%
\patchBothAmsMathEnvironmentsForLineno{alignat}%
\patchBothAmsMathEnvironmentsForLineno{gather}%
\patchBothAmsMathEnvironmentsForLineno{multline}%
}

\begin{document}

%\linenumbers

\title{\huge Dropping Convexity for More Efficient and Scalable Online Multiview Learning}

\date{\today}

\author{Zhehui Chen, Lin F. Yang, Chris J. Li, and Tuo Zhao\thanks{Z. Chen and T. Zhao are affiliated with School of Industrial and Systems Engineering at Georgia Tech; F. L. Yang is affiliated with Department of Computer Science and Department of Physics and Astronomy at Johns Hopkins University; C. J. Li is affiliated with Department of Operations Research and Financial Engineering at Princeton University; Tuo Zhao is the corresponding author; Email:$\{$zhchen, tourzhao$\}$@gatech.edu. A preliminary version is presented at International Conference on Machine Learning, 2017 \citep{chen2017online}.} }

\maketitle
 \begin{abstract}

Multiview representation learning is very popular for latent factor analysis. It naturally arises in many data analysis, machine learning, and information retrieval applications to model dependent structures among multiple data sources.
For computational convenience, existing approaches usually formulate the multiview representation learning as convex optimization problems, where global optima can be obtained by certain algorithms in polynomial time. However, many evidences have corroborated that heuristic nonconvex approaches also have good empirical computational performance and convergence to the global optima, although there is a lack of theoretical justification. Such a gap between theory and practice motivates us to study a nonconvex formulation for multiview representation learning, which can be efficiently solved by a simple stochastic gradient descent (SGD) algorithm. Theoretically, we first illustrate the geometry of the nonconvex formulation; Then by characterizing the dynamics of the algorithm based on diffusion processes, we establish global rates of convergence to the global optima. Numerical experiments are provided to support our theory.
\end{abstract}

%!TEX root = ./draft.tex
\section{Introduction}\label{introduction}
%
%Let $X\in\RR^d$ and $Y\in\RR^m$ be two random vectors following some joint distribution $\cD$. We consider the following optimization problem as follows:
%\begin{align}
%\label{eqn:def}
%(\hat{u},~\hat{v}) = \argmax_{u\in \RR^{d}, v\in\RR^{m}}~\EE~v^\top YX^\top u\quad\textrm{subject~to}~\norm{u}^2 =1, \norm{v}^2 =1.
%\end{align}
%\eqref{eqn:def}

%Existing literature in numerical analysis usually refers \eqref{eqn:def} as the rank one singular value decomposition.

Multiview data have become increasingly available in many popular real-world data analysis and machine learning problems. These data are collected from diverse domains or different feature extractors, which share latent factors. %because single-view data cannot describe the information of all examples well.
%Multiview data are collected can be considered as multiple measurement modalities, obtained from multiple data sources.
Existing literature has demonstrated different scenarios. For instance, the pixels and captions of images can be considered as two-view data, since they are two different features describing the same contents. %Moreover, images in computer vision, text and linguistic data in natural language processing, and speech data in acoustic recognition. 
More motivating examples involving two or more data sets simultaneously can be found in computer vision, natural language processing, and acoustic recognition.  See \cite{hardoon2004canonical, socher2010connecting, kidron2005pixels, chaudhuri2009multi, arora2012kernel, bharadwaj2012multiview, vinokourov2002inferring, NIPS2011_4193}.
Although these data are usually unlabeled, there exist underlying association and dependency between different views, which allows us to learn useful representations in a unsupervised manner. Here we are interested in finding a representation that reveals intrinsic low-dimensional structures and decomposes underlying confounding factors.
One ubiquitous approach is partial least square (PLS) for multiview representation learning. Specifically, given a data set of $n$ samples of two sets of random variables (views), $X\in\RR^{m}$ and $Y\in\RR^{d}$,  PLS aims to find an $r$-dimensional subspace ($r\ll \min(m,d)$) that preserves most of the covariance between two views.
Existing literature has shown that such a subspace is spanned by the leading $r$ components of the singular value decomposition (SVD) of $\Sigma_{XY} = \EE_{(X,Y)\sim \cD}\sbr{XY^\top}$ \cite{arora2012stochastic}, where we sample $(X,Y)$ from some unknown distribution $\cD$. Throughout the rest of the paper, if not clear specified, we denote $\EE_{(X,Y)\sim \cD}$ by $\EE$ for notational simplicity.

A straightforward approach for PLS is ``Sample Average Approximation'' (SAA, \cite{abdi2003partial,ando2005framework}), where we run an offline (batch) SVD algorithm on the empirical covariance matrix after seeing sufficient data samples. However, in the ``big data'' regime, this approach requires unfeasible amount of storage and computation time. Therefore, it is much more practical to consider the multiview learning problem in a ``data laden'' setting, where we draw independent samples from an underlying distribution $\cD$ over $\RR^{m} \times \RR^{d}$, one at a time. This further enables us to formulate PLS as a stochastic (online) optimization problem. Here we only consider the rank-$1$ case ($r=1$) for simplicity, and solve
\begin{align}
\label{eqn:def}
(\hat{u},~\hat{v}) = \argmax_{u\in \RR^{m}, v\in\RR^{d}}~\EE \left(v^\top YX^\top u\right)\quad\textrm{subject~to}\quad u^\top u =1, v^\top v = 1.
\end{align}
%where we rewrite $U$ as a vector $u\in\RR^m$ and $v\in\RR^d$. 
We will explain more details on the rank-$r$ case in the later section. 

Several nonconvex stochastic approximation (SA) algorithms have been proposed in \cite{arora2012stochastic}. 
These algorithms work great in practice, but lack theoretic justifications, since the nonconvex nature of \eqref{eqn:def} makes the theoretical analysis very challenging. 
To overcome this obstacle, \cite{arora2016stochastic} propose a convex relaxation of \eqref{eqn:def}. 
Specifically, by a reparametrization $M = uv^\top$ (Recall that we are interested in the rank-1 PLS), they rewrite \eqref{eqn:def} as\footnote{For $r>1$ case, we replace $\norm{M}_{*}\leq 1$ with $\norm{M}_{*}\leq r$}
\begin{align}\label{PLS-cvx}
\hat{M} = \argmax_{M}~\langle M, \Sigma_{XY} \rangle \quad \textrm{subject to}\quad \norm{M}_{*}\leq 1~\textrm{and}~\norm{M}_2 \leq 1.
\end{align}
where $\Sigma_{XY} = \EE XY^{\top}$, and $\norm{M}_2$ and $\norm{M}_*$ are the spectral (i.e., the largest singular value of $M$) and nuclear (i.e., the sum of all singular values of $M$) norms of $M$ respectively. By examining the KKT conditions of \eqref{PLS-cvx}, one can verify that $\hat{M}=\hat{u}\hat{v}^\top$ is the optimal solution, where $\hat{u}, \hat{v}$ are the leading left and right singular vectors of $\Sigma_{XY}$, i.e., a pair of global optimal solutions to \eqref{eqn:def} for $r=1$. Accordingly, they propose a projected stochastic gradient-type algorithm to solve \eqref{PLS-cvx}, which is often referred to the Matrix Stochastic Gradient (MSG) algorithm. Particularly, at the $(k+1)$-th iteration, MSG takes
\begin{align*}
M_{k+1} = \Pi_{\rm Fantope}(M_k + \eta X_kY_k^\top),
\end{align*}
where $X_k$ and $Y_k$ are independently sampled from $\cD$, and $\Pi_{\rm Fantope}(\cdot)$ is a projection operator to the feasible set of \eqref{PLS-cvx}. They further prove that given a pre-specified accuracy $\epsilon$, MSG requires $N=~\cO(\epsilon^{-2}\log(1/\epsilon))$ iterations such that $\langle \hat{M}, \EE xy^\top \rangle - \langle M_N, \EE xy^\top \rangle \leq \epsilon$ with high probability.

Despite of the attractive theoretic guarantee, MSG does not present superior performance to other heuristic nonconvex stochastic optimization algorithms for solving \eqref{eqn:def}. Although there is a lack of theoretical justification, many evidences have corroborated that heuristic nonconvex approaches not only converge to the global optima in practice, but also enjoy better empirical computational performance than the convex approaches \citep{zhao2015nonconvex,candes2015phase,ge2015escaping,cai2016optimal}. Another drawback of MSG is the complicated projection step at each iteration. Although \cite{arora2016stochastic} further propose an algorithm to compute the projection with a computational cost cubically depending on the rank of the iterates (the worst case: $\cO(d^3)$), such a sophisticated implementation significantly decreases the practicability of MSG. Furthermore, MSG is also unfavored in a memory-restricted scenario, since storing the update  $M^{(k)}$ requires $\cO(md)$ real number storage. In contrast, the heuristic algorithms analyzed in this paper require only $\cO(m + d)$ real number storage, or $\cO(rm+rd)$ in the rank-$r$ case.

%Conventional algorithm that has a theoretic guarantee is the convex relaxed version,
%e.g. \cite{raman}, which is complicated and is hard to be implemented.
%The simple SGD version do have simple form but does not have good theoretic guarantee.

%\paragraph{Stochastic Power Method (SPM):} The PLS objective has gradient $\Sigma_{yx}U$ with respect to $U$ and $\Sigma_{xy}V$ with respect to $V$.
%By written an unbiased estimator of $\Sigma_{yx}$ and $\Sigma_{xy}$, we obtain the updates for the SPM algorithm \cite{arora2012stochastic}.
%\begin{align*}
%U_t &= \Pi_{\text{orth}}\rbr{U_{t-1} + \eta x_ty_t^{T} V_{t-1}} \\
%V_t &= \Pi_{\text{orth}}\rbr{V_{t-1} + \eta y_tx_t^{T} U_{t-1}},
%\end{align*}
%where $\Pi_{\text{orth}}\rbr{\cdot}$ projects the vector onto a unit sphere. 
%The operation of the projection can be usually ignored since it is performed for purely numerical reasons and only very infrequently.
%Therefore, each update step only costs $\cO(m + d)$ operations.

We aims to bridge the gap between theory and practice for solving multiview representation learning problems by nonconvex approaches. Specifically, we first illustrate the nonoconvex geometry of \eqref{eqn:def}, we analyze the convergence properties of a simple stochastic optimization algorithm for solving \eqref{eqn:def} based on diffusion processes. Our analysis takes advantage of the strong Markov properties of the stochastic optimization algorithm updates and casts the trajectories of the algorithms as a diffusion processes \citep{ethier2009markov,li2016online}. By leveraging the weak
convergence from discrete Markov chains to their continuous time limits, we demonstrate that the trajectories are essentially the solutions to stochastic differential equations. Such an SDE-type analysis automatically incorporates the geometry of the objective and the randomness of the algorithm, and eventually demonstrates three phases of convergence.
\begin{enumerate}
\item Starting from an unstable equilibrium with negative curvature, the dynamics of the algorithm can be described by an Ornstein-Uhlenbeck process with a steady driven force pointing away from the initial.

\item When the algorithm is sufficiently distant from the initial unstable equilibrium, the dynamics can be characterized by a deterministic ordinary differential equation (ODE). The trajectory of this phase is evolving directly toward the desired global maximum until it reaches a small basin around the global maximum.

\item In this phase, the trajectory can be also described by an Ornstein-Uhlenbeck process oscillating around the global maximum. The process has a drifting term that gradually dies out and eventually becomes a nearly unbiased random walk centered at the maximum.
\end{enumerate}
The sharp characterization in these three phases eventually allows us to establish strong convergence guarantees. Particularly, we show that the nonconvex stochastic gradient algorithm guarantees an $\epsilon$-optimal solution in $\cO({\epsilon}^{-1}\log(\epsilon^{-1}))$ iterations with high probability, which is a significant improvement over convex MSG by a factor of $\epsilon^{-1}$. Our theoretical analysis reveals the power of the nonconvex optimization in PLS. The simple heuristic algorithms drop the convexity, but achieve much better efficiency.

Our convergence analysis also has important implications on stochastic optimization algorithm for Canonical Correlation Analysis (CCA). Specifically, CCA considers a similar setting to PLS, and solves
\begin{align}\label{CCA-obj}
(\hat{u},\hat{v})=\argmax_{u,v}~u^\top{\EE}XY^\top v\quad\textrm{subject to}\quad  \EE(X^\top u)^2=1,~\EE(Y^\top v)^2=1.
\end{align}
From an optimization perspective, CCA is equivalent to PLS under some linear transformation, but more challenging. We will explain more details on CCA in our later discussions.

%The rest of this paper is organized as follows: In Section \ref{sec:algorithm}, we present two heuristic nonconvex stochastic optimization algorithms; In Sections \ref{ODE} and \ref{SDE}, we present our theoretical analysis on casting the trajectories of the algorithms as a diffusion processes, and establish strong convergence guarantees; In Section \ref{Simulations}, we present numerical experiments to support our theory; In Section \ref{Discussions}, we discuss related work and draw a brief conclusion.

\noindent{\textbf{Notations}}: Given a vector $ v = (v^{(1)}, \ldots, v^{(d)})^\top\in\RR^{d}$, we define vector norms: $\norm{ v}_1 = \sum_j|v^{(j)}|$, $\norm{ v}_2^2 = \sum_j(v^{(j)})^2$, and $\norm{ v}_{\infty} = \max_j|v^{(j)}|$. Given a matrix $A\in\RR^{d \times d}$, we use $A_{j} = (A_{1j},...,A_{dj})^\top$ to denote the $j$-th column of $A$ and define the matrix norms $\norm{A}_{\rm F}^2 = \sum_j\norm{A_{j}}_2^2$ and $\norm{A}_2$ as the largest singular value of $A$.

%!TEX root = ./draft.tex
\section{Stochastic Nonconvex Optimization}\label{sec:algorithm}

Recall that we solve \eqref{eqn:def}
\begin{align}\label{PLS-obj-1}
(\hat{u},\hat{v})=\argmax_{u,v}~u^\top{\EE}XY^\top v\quad\text{subject to}\quad \norm{u}_2^2=1,~\norm{v}_2^2=1,
\end{align}
where $(X,Y)$ follows some unknown distribution $\cD$. Due to the symmetrical structure of \eqref{PLS-obj-1}, $(-\hat{u},-\hat{v})$ is also a pair of global optimum. Our analysis holds for both optima. Throughout the rest of the paper, if not clearly specified, we consider $(\hat{u},\hat{v})$ as the global optimum for simplicity.

We apply the stochastic approximation (SA) of the generalized Hebbian algorithm (GHA) to solve \eqref{PLS-obj-1}. GHA, which is also referred as Sanger's rule \citep{sanger1989optimal}, is essentially a primal-dual algorithm. Specifically, we consider the Lagrangian function of \eqref{PLS-obj-1}:
\begin{align}\label{Lagrangian}
L(u,v,\mu,\sigma) = u^\top{\EE}XY^\top v -&\mu(u^\top u-1) -\sigma(v^\top v-1),
\end{align}
where $\mu$ and $\sigma$ are Lagrangian multipliers. We then check the optimal KKT conditions,
\begin{align}\label{KKT-Cond}
{\EE}XY^\top v - &2\mu u = 0,\quad{\EE}YX^\top u - 2\sigma v = 0,\quad u^\top u = 1\quad\textrm{and}\quad  v^\top v = 1,
\end{align}
which further imply
\begin{align*}
&u^\top\EE XY^\top v - 2\mu u^\top u = u^\top\EE XY^\top v-2\mu=0,\\
&v^\top{\EE}YX^\top u - 2\sigma v^\top v = v^\top{\EE}YX^\top u - 2\sigma= 0.
\end{align*}
Solving the above equations, we obtain the optimal Lagrangian multipliers as 
\begin{align}\label{optimal-multiplier}
\mu = \sigma = \frac{1}{2}u^\top\EE XY^\top v.
\end{align}
GHA is inspired by \eqref{KKT-Cond} and \eqref{optimal-multiplier}. At $k$-th iteration GHA takes 
\begin{align}\label{eqn:Dual}
\textrm{Dual  Update :}\quad \mu_k ~~= ~~\sigma_k ~~=~~ \frac{1}{2}\hspace{-0.625in}\underbrace{u_k^\top X_kY_k^\top v_k}_{\textrm{SA (stochastic approximation) of $u_k^\top \Sigma v_k$}}\hspace{-0.625in},
\end{align}
\begin{align}\label{eqn:Primal}
\textrm{Primal Update :}\quad u_{k+1}~~=~~u_k+\eta\underbrace{\left(X_kY_k^\top v_k -2\mu_k u_k \right)}_{\textrm{SA of $\nabla_u L(u,v,\mu,\sigma)$} },\quad v_{k+1}~~=~~v_k+\eta\underbrace{\left(Y_kX_k^\top u_k -2\sigma_k v_k \right)}_{\textrm{SA of $\nabla_v L(u,v,\mu,\sigma)$} },
\end{align}
where $\eta>0$ is the step size. Combining \eqref{eqn:Dual} and \eqref{eqn:Primal}, we obtain a dual-free update as follow:
\begin{align}
u_{k+1}~~=~~u_k+\eta\left(X_kY_k^\top v_k -u_k^\top X_kY_k^\top v_k  u_k \right) \quad \textrm{and}\quad v_{k+1}~~=~~v_k+\eta\left(Y_kX_k^\top u_k -u_k^\top X_kY_k^\top v_k v_k \right).\label{free-update}
\end{align}
Different from the projected SGD algorithm, which is a primal algorithm proposed in \cite{chen2017online}, Stochastic GHA does not need projection at each iteration.
%In the following sections, we first define the stable and unstable stationary point. Then we use ODE and SDE to characterize the trajectory of our algorithm. Last, we present some experiments to support our analysis.

%!TEX root = ./draft.tex
\section{Optimization Landscape}\label{CSS}

We illustrate the nonconvex optimization landscape of \eqref{eqn:def}, which helps us understand the intuition behind the algorithmic convergence. We first study its stationary points based the Lagrangian function \eqref{Lagrangian}. By the KKT conditions \eqref{KKT-Cond}, we define the stationary point of \eqref{Lagrangian} as follows.
\begin{definition}\label{def-stationary}
Given \eqref{eqn:def} and \eqref{Lagrangian}, we define:
\begin{enumerate}
\item A quadruplet of $(u,v,\mu,\sigma)$ is called a stationary point of \eqref{Lagrangian}, if it satisfies \eqref{KKT-Cond}. 
\item A pair of $(u,v)$ is called a stable stationary point of ~\eqref{eqn:def}, if $(u,v,\mu,\sigma)$ is a stationary point of \eqref{Lagrangian}, and $\nabla_{u,v}^2 L(u,v,\mu,\sigma)$ is negative semi-definite.
\item A pair of $(u,v)$ is called an unstable stationary point of \eqref{eqn:def}, if $(u,v,\mu,\sigma)$ is a stationary point of \eqref{Lagrangian}, and $\nabla_{u,v}^2 L(u,v,\mu,\sigma)$ has a positive eigenvalue.
\end{enumerate}
\end{definition}
%Definition \ref{def-stationary} defines two types of stationary points: One is the maximum (including local and global) of the  \eqref{eqn:def}, and the other is the unstable stationary point.
We then obtain all stationary points by solving \eqref{KKT-Cond}. For notational simplicity, we denote $\Sigma_{XY} = \EE XY^\top$. Before we proceed with our analysis, we introduce the following assumption.
\begin{assumption}\label{mild}
Suppose $d\leq m$ and rank$(\Sigma_{XY})=r$. We have $\lambda_1 >\lambda_2\geq \lambda_3\geq\cdots\geq\lambda_r>0$, where $\lambda_i$'s are the $i$-th singular values of $\Sigma_{XY}$.
\end{assumption}
We impose such an eigengap assumption ($\lambda_1 >\lambda_2$) to ensure the identifiability of the leading pair of singular vectors. Thus, the leading pair of singular vectors are uniquely determined only up to sign change.
%First, we find out all the stationary points of \eqref{eqn:def}.
Let $O_1\in \RR^{m\times m}$ and $O_2\in \RR^{d \times d}$ be any pair of left and right singular matrices\footnote{Since all singular values are not necessarily distinct, some pairs of singular vectors are not unique, e.g., when $\lambda_i=\lambda_j$, $(\overline{u}_i,\overline{v}_i)$ and $(\overline{u}_j,\overline{v}_j)$ are uniquely determined up to rotation. Note that our analysis works for all possible combinations of $O_1$ and $O_2$. See more details in \cite{golub2012matrix}.}. Let $\overline{u}_i$ and $\overline{v}_j$ denote the $i$-th column of $O_1$ and $j$-th column of $O_2$, respectively. The next proposition reveals the connection between stationary points and singular vectors.
\begin{proposition}\label{geometry}
Suppose Assumption~\ref{mild} holds. A quadruplet  of $(u,v,\mu,\sigma)$ is the stationary point of \eqref{Lagrangian}, if either of the following condition holds:
\begin{enumerate}
\item $(u,v)$ are a pair of singular vectors associated with the same nonzero singular value;
\item $u$ and $v$ belong to the row and column null spaces of $\Sigma_{XY}$ respectively: $\Sigma_{XY} v=0,~\Sigma_{XY}^\top u=0$.
\end{enumerate}
\end{proposition}
The proof of Proposition \ref{geometry} is presented in Appendix \ref{geometry-proof}. We then determine the types of these obtained stationary points. The next proposition characterizes the maximum eigenvalues of $\nabla_{u,v}^2 L(u,v,\mu,\sigma)$ at these stationary points of \eqref{Lagrangian}.

\begin{proposition}\label{stable}
Suppose Assumption~\ref{mild} holds. All pairs of singular vectors associated with the leading singular value are global optima of \eqref{eqn:def}, i.e., also the saddle points of \eqref{Lagrangian}, and they are stable stationary points. All other stationary points of \eqref{Lagrangian} are all unstable with
\begin{align*}
\lambda_{\max}(\nabla_{u,v}^2 L(u,v,\mu,\sigma))\geq \lambda_1-\lambda_2.
\end{align*}
\end{proposition}
The proof of Proposition~\ref{stable} is presented in Appendix \ref{stable-proof}. Proposition~\ref{stable} essentially  characterizes the geometry of \eqref{eqn:def} at all stationary points, and the unstableness allows the stochastic gradient algorithm to escape, as will be shown in the next sections.
 %Since we already have the geometry of \eqref{eqn:def}, now we can character the convergence of our algorithms.

%Therefore, we know $(u^\top A v)^2$ is an eigen-value of $A^\top A$ as well as $AA^\top$, assume it equals to $l_k^2$, $k\in \{1,2,\cdots,d\}$. Then $u$ is the eigen-vector of $AA^\top$ corresponding to $l_k^2$, i.e., $u$ is the $k$-th column of $O_1$ or its opposite direction vector. Similarly, we know that $v$ is the $k$-th column of $O_2$ or its opposite direction vector. Given all $l_i$, where $i=1,2,...,d$, are different, $u$ and $v$ are unique. Hence, $ \left(\frac{1}{u^\top A v}A^\top A-u^\top A v \cdot I_d \right)\sim \frac{1}{u^\top A v}\diag(l_1^2-l_k^2,\cdots,l_d^2-l_k^2).$ As we can see, the hessian matrix is negative semi-definite(original problem is a maximizing problem) if and only if $u$ is the first column of $O_1$ (opposite direction) and $v$ is the first column of $O_2$ (opposite direction). We need $u^\top A v>0$, because the left upper part and the right lower part both need to be negative. In this case the stationary points are the optima, i.e., stable stationary points. Otherwise, the Hessian either has a negative curvature or is positive semi-definite, which means these stationary points are not stable. 
%
%By the above analysis, we know in problem~\eqref{eqn:def}, there are only two maximizers. We already know the geometry of the problem~\eqref{eqn:def}. In the following section, we show the global convergence of algorithm~\eqref{PSG-update-u-v}.

%!TEX root = ./draft.tex
\section{Global Convergence by ODE}\label{ODE}

Before we proceed with our analysis, we first impose some mild assumptions on the problem.

\begin{assumption}\label{Assumption1}

$X_k,Y_k,~k=1,2,...N$ are data samples identically independently distributed as $X \in \RR^d,~Y \in \RR^d$ respectively satisfying the following conditions: 
\begin{enumerate}
\item For any $\Delta>0$, $\max\{\EE\norm{X}_2^{4+\Delta},\EE\norm{Y}_2^{4+\Delta}\}<\infty$ and $\max\{\EE\norm{X}_2^2,\EE\norm{Y}_2^2\}\leq Bd$ for a constant $B$;\footnote{We only need ($4+\Delta$)-th moments of $\norm{X}_2$ and $\norm{Y}_2$ to be bounded, while the preliminary results in \cite{chen2017online} require both $\norm{X}_2$ and $\norm{Y}_2$ to be bounded random variables.} 
%\item $\norm{X}_2^2\leq B, ~\norm{Y}_2^2\leq B$ for a constant $B$; 
\item $ \lambda_1 >\lambda_2\geq  \lambda_3\geq ...\geq \lambda_d>0$, where $\lambda_i$'s are the singular values of $\Sigma_{XY}=\EE XY^\top$.
\end{enumerate}
\end{assumption}

%For simplicity, we assume that $X$ and $Y$ are bounded random vectors, but the extension to sub-Gaussian random vectors with $B\asymp d$ is straightforward.
Here we assume $X$ and $Y$ are of the same dimensions (i.e., $m=d$) and $\Sigma_{XY}$ is full rank for convenience of analysis. The extension to $m\neq d$ in a rank deficient setting is straightforward, but more involved (See more details in Section \ref{mnotequald}). Moreover, for a multiview learning problem, it is also natural to impose the following additional assumptions.

\begin{assumption}\label{Assumption2}
Given the observed random variables $X$ and $Y$, there exist two orthogonal matrices $O_X \in \RR^{d\times d},~O_Y \in \RR^{d\times d}$ such that $X=O_X \overline{X},~Y=O_Y \overline{Y}$, where $\overline{X}=(\overline{X}^{(1)},...,\overline{X}^{(d)})^\top\in\RR^d$ and $\overline{Y}=(\overline{Y}^{(1)},...,\overline{Y}^{(d)})^\top\in\RR^{d}$ are the latent variables satisfying:
\begin{enumerate}
\item $\overline{X}^{(i)}$ and $\overline{Y}^{(j)}$ are uncorrelated if $i \neq j$, so that $O_X$ and $O_Y$ are the left and right singular matrices of $\Sigma_{XY}$ respectively;
\item $\Var(\overline{X}^{(i)})=\gamma_i,~~\Var(\overline{Y}^{(i)})=\omega_i,~~\EE \left(\overline{X}^{(i)}\overline{Y}^{(i)}\overline{X}^{(j)}\overline{Y}^{(j)}\right)=\alpha_{ij}$, where $\gamma_i,\alpha_{ij}$, and $\omega_i$ are constants.
\end{enumerate}
\end{assumption}

The next proposition characterizes the strong Markov property of our algorithm.

\begin{proposition}\label{Markov}
Using \eqref{free-update}, we get a sequence of $(u_k,v_k),~k=1,2,..., N$. They form a discrete-time Markov process. % Moreover, for  \eqref{PSG-update-u-v}, each $u_k$ and $v_k$ are on the $\cS^{d-1}$.
\end{proposition}

With Proposition \ref{Markov}, we can construct a continuous time process to derive an ordinary differential equation to analyze the algorithmic convergence. Specifically, as the fixed step size $\eta \rightarrow 0^+$, two processes $U_\eta(t)=u_{\lfloor \eta^{-1}t\rfloor}, ~V_\eta(t)=v_{\lfloor \eta^{-1}t\rfloor}$ based on the sequence generated by \eqref{free-update}, weakly converge to the solution of the following ODE system in probability (see more details in \cite{ethier2009markov}),

\begin{align}
\frac{dU}{dt}=\left( \Sigma_{XY} V - U^\top \Sigma_{XY} V U \right),\quad \frac{dV}{dt}=\left( \Sigma_{XY}^\top U - V^\top \Sigma_{XY}^\top U V \right),\label{limitation u-v}
\end{align}
where $U(0)=u_0$ and $V(0)=v_0$.
To highlight the sequence generated by \eqref{free-update} depending on $\eta$, we redefine $u_{\eta,k}=u_k, ~v_{\eta,k}=v_k$.

\begin{theorem}\label{Converge-ODE}
%If $(u_{\eta,0},v_{\eta,0})$ converges weakly to some constant vector $U_0,~V_0$, then 
As $\eta \rightarrow 0^+$, the processes $u_{\eta,k},~v_{\eta,k}$ weakly converge to the solution of the ODE system in \eqref{limitation u-v} % and \eqref{limitation v} 
with sphere initial $U(0)=u_0,~V(0)=v_0$, i.e., $\norm{u_0}_2=\norm{v_0}_2=1$. 
\end{theorem}
The proof of Theorem \ref{Converge-ODE} is presented in Appendix~\ref{Converge-ODE-proof}. Under Assumption~\ref{Assumption1}, the above ODE system admits a closed form solution. Specifically, we solve $U$ and $V$ simultaneously, since they are coupled together in \eqref{limitation u-v}% and \eqref{limitation v}
. To simplify \eqref{limitation u-v}% and \eqref{limitation v}
, we define $W=\frac{1}{\sqrt{2}}\left( U^\top~ V^\top \right)^\top $ and $w_k=\frac{1}{\sqrt{2}}\left(u_k^\top~v_k^\top\right)^\top$. We then rewrite \eqref{limitation u-v}% and \eqref{limitation v}
 as
\begin{align}\label{To-Simplify}
&\frac{dW}{dt}=QW-W^\top Q W W,
\end{align}
where $Q=\left(
\begin{array}{cc}
0 & \Sigma_{XY}\\
\Sigma_{XY}^\top & 0
\end{array}
\right)$.
By Assumption~\ref{Assumption2}, $O_X$ and $O_Y$ are the left and right singular matrices of $\Sigma_{XY}$ respectively, i.e., $\Sigma_{XY}=\EE XY^\top=O_X\EE \overline{X}\overline{Y}^\top O_Y^\top$, where $\EE \overline{X}\overline{Y}^\top$ is diagonal. For notational simplicity, we define $D=\diag(\lambda_1,\lambda_2,...,\lambda_d)$
such that $\Sigma_{XY}=O_X D O_Y^\top$. One can verify $Q=P \Lambda P^\top$, where
\begin{align}\label{Trans}
P=\frac{1}{\sqrt{2}}\left(
\begin{array}{cc}
O_X & O_X\\
O_Y &-O_Y
\end{array}
\right)\quad\textrm{and}\quad\Lambda=\left(
\begin{array}{cc}
D & 0\\
0 & -D
\end{array}
\right).
\end{align}
By left multiplying $P^\top$ both sides of \eqref{To-Simplify}, we obtain
\begin{align}\label{Simplified1}
H(t)=P^\top W(t)~\textrm{with}~\frac{dH}{dt}=\Lambda H-H^\top \Lambda HH,
\end{align}
which is a coordinate separable ODE system. Accordingly, we define $h_k^{(i)}$'s as: 
\begin{align}\label{transformed_solution_w}
h_k=P^\top w_k \quad \textrm{and}\quad h_k^{(i)}=P_i^\top w_k.
\end{align}
Thus, we can obtain a closed form solution to \eqref{Simplified1} based on the following theorem.
\begin{theorem}\label{Solution-ODE}
Given \eqref{Simplified1}, we write the ODE in each component $H^{(i)}$,
\begin{align}\label{Simplified2}
\frac{d}{dt}H^{(i)}=H^{(i)}\sum_{j=1}^{2d}\left(\lambda_i-\lambda_j\right)(H^{(j)})^2,
\end{align}
where $\lambda_i=-\lambda_{i-d}$ when $i>d$.
This ODE System has a closed form solution as follows:
\begin{align}\label{solution}
 H^{(i)}(t)=\big(C(t) \big)^{-\frac{1}{2}}H^{(i)}(0)\exp{(\lambda_i t)},
\end{align}
for $i=1,2,...,2d$, where $$C(t)=\sum\limits_{j=1}^{2d}\left(\left(H^{(j)}(0)\right)^2\exp{(2\lambda_j t)}\right)$$ is a normalization function such that $\norm{H(t)}_2=1$.
\end{theorem}

%
%\begin{theorem}
%Solution of \eqref{final} is $H^{(i)}(t)=\left(C(t)\right)^{-\frac{1}{2}}H^{(i)}(0)\exp{(\lambda_i t)}$,
%%\begin{align}\label{Sol of ODE}
%%H^{(i)}(t)=\left(C(t)\right)^{-\frac{1}{2}}H^{(i)}(0)\exp{(\lambda_i t)},
%%\end{align}
%where $C(t)=\sum\limits_{j=1}^{2d}\left(H_j^2(0)\exp{(2\lambda_j t)}\right)$ is a normalized parameter, which guarantees $\norm{H(t)}=1$.\\
%\end{theorem}
%The solution of Eq.\eqref{final} is 
%\begin{align}\label{Sol of ODE}
%H_i(t)=\left(C(t)\right)^{-\frac{1}{2}}H_i(0)\exp{(\lambda_i t)}
%\end{align}
%where $C(t)=\sum\limits_{j=1}^{2d}\left(H_j^2(0)\exp{(2\lambda_j t)}\right)$ is a normalized parameter, which guarantees $\norm{H(t)}=1$\\

The proof of Theorem \ref{Solution-ODE} is presented in Appendix \ref{Solution-ODE-proof}. Without loss of generality, we assume $H^{(1)}(0)>0$. As can be seen, $H_1(t) \rightarrow 1$, as $t\rightarrow \infty$. We have successfully characterized the global convergence performance of our algorithm with an approximate error $o(1)$. The solution to the ODE system in \eqref{solution}, however, does not fully reveal the algorithmic behavior (more precisely, the rate of convergence) near the equilibria of the ODE system. This further motivates us to exploit the stochastic differential equation approach to characterize the dynamics of the algorithm.

\section{Global Dynamics by SDE}\label{SDE}
%\vspace{-0.05in}

We analyze the dynamics of the algorithm near the equilibria based on stochastic differential equation by rescaling analysis. Specifically, we characterize three stages for the trajectories of solutions: [a] Neighborhood around unstable equilibria --- minimizers and saddle points of \eqref{PLS-obj-1}, [b] Neighborhood around stable equilibria --- maximizers of \eqref{PLS-obj-1}, and [c] deterministic traverses between equilibria. Moreover, we provide the approximate the number of iterations in each phase until convergence.

%one is stable which is the optimal point, the other is unstable, which is other stationary points, however, in PLS problem, the stable region for other stationary points is too small and with a random start point, it is unlikely for the algorithm to fall in to these regions. Also, we find that the theoretical analysis of these two scenarios are exactly similar.

%Firstly, we proof that this algorithm iterations can get rid of the unstable equilibria(including all eigenvectors except the largest absolute ).

%\textbf{TO CHANGE THE PROVE ORDER}

%\vspace{-0.15in}
\subsection{Phase I: Escaping from Unstable Equilibria}
%\vspace{-0.05in}

Suppose that the algorithm starts to iterate around a unstable equilibrium, (e.g. saddle point). Different from our previous analysis, we rescale two aforementioned processes $U_{\eta}(t)$ and $V_{\eta}(t)$ rescaled by a factor of $\eta^{-1/2}$. This eventually allows us to capture the uncertainty of the algorithm updates by stochastic differential equations. Roughly speaking, the ODE approximation is essentially a variant of law of large number for Markov process, while the SDE approximation serves as a variant of central limit theorem accordingly. 

%We assume that the starting point is $A_j$, which is the $j^{th}$ column of A. By calculating the rescaled we can get its approximate Stochastic Differential Equation. With it, we can find how it can escape from the unstable equilibria stage. As we mentioned before, we need rescale the process, let 

Recall that $P$ is an orthonormal matrix for diagonalizing Q, and $H$ is defined in~\eqref{Simplified1}. Let $Z_\eta^{(i)}$ and $z_{\eta,k}^{(i)}$ denote the $i$-th coordinates of $Z_\eta=\eta^{-1/2}H_\eta$ and $z_{\eta,k}=\eta^{-1/2}h_{\eta,k}$ respectively. The following theorem characterizes the dynamics of the algorithm around the unstable equilibrium.
% In fact we can get a simular formula just like \ref{final} and \ref{6}, the difference is $h\approx e_j$. 
%By calculating the infinitesimal conditional expectation and infinitesimal conditional variance we can get its approximate Stochastic Differential Equation. With it, we can find how it can escape from the unstable equilibria stage. The calculation is just similar like \ref{4} and \ref{5}\
%\vspace{-0.025in}
\begin{theorem}\label{SDE for saddle}
Suppose $z_{\eta,0}$ is initialized around some saddle point or minimizer (e.g. $j$-th column of $P$ with $j\neq 1$), i.e., $Z^{(j)}(0) \approx \eta^{-\frac{1}{2}}$ and $Z^{(i)}(0) \approx 0$ for $i\neq j$. Then for any $C>0,$ there exist $\tau>0$ and $\eta'>0$ such that
\begin{align}\label{sde:con}
\sup_{ \eta<\eta'}\PP(\sup_t |Z_{\eta}^{(i)}(t)|\leq C)\leq 1-\tau.
\end{align}

%\begin{align}\label{diffusion-SDE}
%d Z^{(i)}(t)=-(\lambda_j-\lambda_i)Z^{(i)}(t)dt+\beta_{ij}dB(t), 
%\end{align}
%where $B(t)$ is a brownian motion, and $\beta_{ij}$ is defined as follow:
%
%\begin{align*}
%\beta_{ij}=\left\{
%\begin{array}{ll}
%\displaystyle\frac{1}{2}\sqrt{\gamma_i\omega_j+\gamma_j\omega_i+2\alpha_{ij}} &\textrm{if}~1\leq i,j \leq d~\textrm{or}~d+1\leq i,j \leq 2d,\\
%\displaystyle\frac{1}{2}\sqrt{\gamma_i\omega_j+\gamma_j\omega_i-2\alpha_{ij}} &\textrm{otherwise},
%\end{array}
%\right.
%\end{align*}
%where $\gamma_i=\gamma_{i-d}~\textrm{for}~i>d$, $\omega_j=\omega_{j-d}~\textrm{for}~j>d$, similar definition of $\alpha_{ij}$ for $i>d$ or $j>d$.
\end{theorem}
Here we provide the proof sketch and leave the whole proof of Theorem \ref{SDE for saddle} in Appendix \ref{SDE for saddle-proof}. 
\begin{proof}[Proof Sketch]
 We prove this argument by contradiction. Assume the conclusion does not hold, that is there exists a constant $C>0,$ such that for any  $\eta'>0$ we have $$\sup_{\eta\leq \eta'}\PP(\sup_t |Z_{\eta}^{(i)}(t)|\leq C)=1.$$ That implies there exists a sequence $\{\eta_n\}_{n=1}^\infty$ converging to $0$ such that 
\begin{align}\label{eq_contra}
\lim_{n\rightarrow\infty}\PP(\sup_t |Z_{\eta_n}^{(i)}(t)|\leq C)= 1.
\end{align}
Then we show  $\{Z^{(i)}_{\eta_n}(\cdot)\}_n$ is tight and thus converges weakly. Furthermore, $\{Z^{(i)}_{\eta_n}(\cdot)\}_n$ weakly converges to a stochastic differential equation,
\begin{equation}\label{diffusion-SDE}
dZ^{(i)}(t)=-(\lambda_j-\lambda_i)Z^{(i)}(t)dt+\beta_{ij}dB(t).
\end{equation}
We compute the solution of this stochastic differential equation and then show \eqref{sde:con} does not hold. 
\end{proof}
Theorem \ref{SDE for saddle} implies that for $i > j$, with a constant probability $\tau$, escapes from the saddle points at some time $T_1$, i.e., $(H^{(j)}(T_1))^2$ is smaller than $1-\delta^2$, where $(\delta = O(\sqrt{\eta}))$. Note that \eqref{diffusion-SDE} is a Fokker-Planck equation, which admits a closed form solution as follows,
\begin{align}\label{Diffusion-Solution}
Z^{(i)}(t) & = Z^{(i)}(0)\exp\left[-(\lambda_j-\lambda_i)t\right]+\beta_{ij}\int_0^t\exp\left[(\lambda_j-\lambda_i)(s-t)\right]dB(s) \notag\\
& = \Bigg[\underbrace{Z^{(i)}(0)+\beta_{ij}\int_0^t\exp\left[(\lambda_j-\lambda_i)s\right]dB(s)}_{T_1}\Bigg]\underbrace{\exp\left[(\lambda_i-\lambda_j)t\right]}_{T_2}\quad\textrm{for}~i\neq j.
\end{align}
Such a solution is well known as the Ornstein-Uhlenbeck process \citep{oksendal2003stochastic}, and also implies that the distribution of $z^{(i)}_{\eta,k}$ can be well approximated by the normal distribution of $Z^{(i)}(t)$ for a sufficiently small step size. This continuous approximation further has the following implications:
\begin{enumerate}
\item [{\bf [a]}] For $\lambda_i>\lambda_j,~T_1=\beta_{ij}\int_0^t\exp\big[(\lambda_j-\lambda_i)s\big]dB(s)+Z^{(i)}(0)$ is essentially a random variable with mean $Z^{(i)}(0)$ and variance smaller than $\frac{\beta_{ij}^2}{2(\lambda_i-\lambda_j)}$. The larger $t$ is, the closer its variance gets to this upper bound. While $T_2=\exp\big[(\lambda_i-\lambda_j)t\big]$ essentially amplifies $T_1$ by a factor exponentially increasing in $t$. This tremendous amplification forces $Z^{(i)}(t)$ to quickly get away from $0$, as $t$ increases.

\item [{\bf [b]}] For $\lambda_i<\lambda_j$, we have
\begin{align*}
&\EE [Z^{(i)}(t)] = Z^{(i)}(0)\exp\left[-(\lambda_j-\lambda_i)t\right]\quad\textrm{and}\quad\Var [Z^{(i)}(t)]=\frac{\beta_{ij}^2}{2(\lambda_j-\lambda_i)}\big[1-\exp\left[-2(\lambda_j-\lambda_i)t\right]\big].
\end{align*}
As has been shown in {\bf [a]} that $t$ does not need to be large for $Z^{(i)}(t)$ to get away from $0$. Here we only consider relatively small $t$. Since the initial drift for $Z^{(i)}(0)\approx 0$ is very small, $Z^{(i)}$ tends to stay at $0$. As $t$ increases, the exponential decay term makes the drift quickly become negligible. Moreover, by mean value theorem, we know that the variance is bounded, and increases far slower than the variance in {\bf [a]}. Thus, roughly speaking, $Z^{(i)}(t)$ oscillates near $0$.

\item [{\bf [c]}] For $\lambda_j=\lambda_i$, we have $\EE[Z^{(i)}(t)]=Z^{(i)}(0)$ and $\Var [Z^{(i)}(t)] = \beta_{ij}^2$. This implies that $Z^{(i)}(t)$ also tends to oscillate around $0$, as $t$ increases.
\end{enumerate}
Overall speaking, {\bf [a]} is dominative so that it is the major driving force for the algorithm to escape from this unstable equilibrium. More precisely, let us consider one special case for Phase I, that is we start from the second maximum singular value, with $h_{\eta,k}^{(2)}(0)=1$. We then approximately calculate the number of iterations to escape Phase I using the algorithmic behavior of $h_{\eta,k}^{(1)} = \eta^{1/2} z_{\eta,k}^{(1)}\approx  \eta^{1/2} Z_\eta^{(1)}(t)$ with $t=k\eta$ by the following proposition.

%\vspace{-0.025in}
\begin{proposition}\label{Time_Saddle}
Given pre-specified $\nu>0$ and sufficiently small $\eta$, there exists some $\delta\asymp\eta^{\mu}$, where $\mu\in(0,0.5)$ is a generic constant, such that the following result holds: We need at most
\begin{align*}
N_1=\frac{\eta^{-1}}{2(\lambda_1-\lambda_2)}\log\left(\frac{2\eta^{-1}\delta^2(\lambda_1-\lambda_2)}{\Phi^{-1}\left(\frac{1+\nu}{2}\right)^2\beta^2_{12}} +1\right)
\end{align*}
iterations such that $(h_{\eta,N_1}^{(2)})^2\leq 1-\delta^2$ with probability at least $1-\nu$, where $\Phi(x)$ is the CDF of standard normal distribution.
\end{proposition}
%\vspace{-0.025in}

The proof of Proposition \ref{Time_Saddle} is provided in Appendix \ref{Time_Saddle-proof}. Proposition \ref{Time_Saddle} suggests that SGD can escape from unstable equilibria within a few iterations. After escaping from the saddle, SGD gets into the next phase, which is a deterministic traverse between equilibria. 
%\begin{align}\label{Diffusion-Solution}
%Z^{(i)}(t) & = Z^{(i)}(0)\exp\left(-(\lambda_j-\lambda_i)t\right)+\beta_{ij}\int_0^t\exp\left((\lambda_j-\lambda_i)(s-t)\right)dB(s) \notag\\
%& = \left[ Z^{(i)}(0)+\beta_{ij}\int_0^t\exp\left((\lambda_j-\lambda_i)s\right)dB(s) \right] \exp\left((\lambda_i-\lambda_j)t\right),\qquad  i\neq j.
%\end{align}
%Such a solution is well known as the Ornstein-Uhlenbeck process with the following implications.
%\begin{description}
%\item (1) For $\lambda_i<\lambda_j$, we have
%\begin{align*}
%\EE [Z^{(i)}(t)] = Z^{(i)}(0)\exp\left(-(\lambda_j-\lambda_i)t\right)=0\quad\textrm{and}\quad \Var [Z^{(i)}(t)]=\frac{\beta_{ij}^2}{2(\lambda_j-\lambda_i)}\left[1-\exp\left(-2(\lambda_j-\lambda_i)t\right)\right],
%\end{align*}
%This implies that $Z^{(i)}(t)$ tends to oscillate around $0$, as $t\rightarrow\infty$.
%
%\item (2) For $\lambda_j=\lambda_i$, we have $\EE[Z^{(i)}(t)]=0$ and $\Var [Z^{(i)}(t)] = \beta_{ij}^2$. This implies that $Z^{(i)}(t)$ also tends to oscillate around $0$, as $t\rightarrow\infty$.
%
%\item (3) For $\lambda_j<\lambda_i$. We separate $Z^{(i)}(t)$ as two parts. The first part is $Z^{(i)}(0)+\beta_{ij}\int_0^t\exp\left((\lambda_j-\lambda_i)s\right)dB(s)$, which is a random variable with mean $Z^{(i)}(0)$ and variance less than $\frac{\beta_{ij}^2}{2(\lambda_i-\lambda_j)}$. The larger $t$ is, the closer its variance is to this bound. And the next part is $\exp\left((\lambda_i-\lambda_j)t\right)$, which is an exponential increasing part.
%\end{description}

%\vspace{-0.125in}
\subsection{Phase II: Traverse between Equilibria}
%\vspace{-0.05in}

When the algorithm is close to neither the saddle points nor the optima, the algorithm's performance is nearly deterministic. Since $Z(t)$ is a rescaled version of $H(t)$, their trajectories are similar. Like before, we have the following proposition to calculate the approximate iterations, $N_2$, following our results in Section~\ref{ODE}. We restart the counter of iteration by Proposition~\ref{Markov}.
%Specifically, we construct $R_k^{(i)}=\frac{Z_k^{(i)}}{Z_k^{(1)}},~i=2,3,...,2d$, where $Z_k^{(i)}$ is the $i$-th component of $Z_k=\eta^{-\frac{1}{2}}h_k=\eta^{-\frac{1}{2}}P^\top w_k$, then
%\begin{align}\label{Rate}
%R_k^{(i)} \approx \frac{H^{(i)}(t)}{H^{(1)}(t)}=\frac{\left(C(t)\right)^{-1/2}H^{(i)}(0)\exp{(\lambda_i t)}}{\left(C(t)\right)^{-1/2}H^{(1)}(0)\exp{(\lambda_1t)}}=R_0^{(i)}\exp{\left(\left(\lambda_k-\lambda_1\right)\eta k\right)},
%\end{align}
%where $t=k\eta$. In other words, $u_k^{(j)}$ is approximately a deterministic curve that decays exponentially to 0 at the rate of $\eta(\lambda_1-\lambda_j)$. This provides us the insight of ODE approximation which guides our convergence analysis of algorithm.

%\vspace{-0.025in}
\begin{proposition}\label{Time_Deterministic}
After restarting counter of iteration, given sufficiently small $\eta$ and $\delta$ defined in Proposition \ref{Time_Saddle}, we need at most $$N_2=\frac{\eta^{-1}}{2(\lambda_1-\lambda_2)}\log\frac{1-\delta^2}{\delta^2}$$ iterations such that $\left(h_{\eta,N_2}^{(1)} \right)^2\geq 1-\delta^2$.
\end{proposition}
The proof of Proposition \ref{Time_Deterministic} is provided in Appendix \ref{Time_Deterministic-proof}. Combining Propositions \ref{Time_Saddle} and \ref{Time_Deterministic}, we know that after $N_1+N_2$ iteration numbers, SGD is close to the optimum with high probability, and gets into its third phase, i.e., convergence to stable equilibria.
%\vspace{-0.025in}

%From \eqref{Rate}, we know that for each component k, it is decreasing exponentially with a rate $\eta(\lambda_1-\lambda_k)$ .
%
%Therefore, in fact when the point is not a eigenvector, the algorithm will make the inner product between point in each iteration and the optimal change very fast. In this case, in fact we can say the phase \Rmnum{1} does not exist, since it is too unstable.
%\vspace{-0.125in}
\subsection{Phase~III:~Convergence~to~Stable~Equilibria}
%\vspace{-0.05in}
%From \eqref{solution} we know that there exists such a constant $T_2$, which only depend on $\delta \in(0,1/2]$, that when $t \geq T_2$, $H_1(t)\geq1-\delta$. This means after $T_2$ times of phase \Rmnum{2}, the algorithm is in can get close enough to the optimal, if $\eta$ is small enough.
% %Thus, for $Z^{(i)}_\eta(t)=\eta^{-1/2}h_{\lfloor \eta^{-1}t\rfloor}^k$,  $k=1,...,2d$\\
%
Again, we restart the counter of iteration by the strong Markov property. The trajectory and analysis are similar to Phase I, since we also characterize the convergence using an Ornstein-Uhlenbeck process. The following theorem characterizes the dynamics of the algorithm around the stable equilibrium.

%\vspace{-0.025in}
\begin{theorem}\label{Convergence}
Suppose $z_{\eta,0}$ is initialized around some maximizer (the first column of $P$), i.e., $Z^{(1)}(0) \approx \eta^{-\frac{1}{2}}$ and $Z^{(i)}(0) \approx 0$ for $i\neq 1$. Then as $\eta \rightarrow 0^+$, for all $i\neq 1$, $z_{\eta,k}^{(i)}$ weakly converges to a diffusion process $Z^{(i)}(t)$ satisfying the following SDE for  $i\neq 1$,
\begin{align}\label{Converge-Solution}
dZ^{(i)}(t)=-(\lambda_1-\lambda_i)Z^{(i)}(t)dt+\beta_{i1}dB(t),
\end{align}
where $B(t)$ is a brownian motion, and 
\begin{align*}
\beta_{i1}=\left\{
\begin{array}{ll}
\displaystyle\frac{1}{2}\sqrt{\gamma_i\omega_1+\gamma_1\omega_i+2\alpha_{i1}} &\textrm{if}~1\leq i \leq d,\\
\displaystyle\frac{1}{2}\sqrt{\gamma_i\omega_1+\gamma_1\omega_i-2\alpha_{i1}} &\textrm{otherwise}.
\end{array}
\right.
\end{align*}
%\textbf{HERE NEED CHANGE THE COEFFICIENT OF BROWNIAN MOTION}
\end{theorem}
%\vspace{-0.025in}

The proof of Theorem \ref{Convergence} is provided in Appendix \ref{Convergence-proof}. Similar to \eqref{Diffusion-Solution}, the closed form solution to \eqref{Converge-Solution} for $i\neq 1$ is as follow:
\begin{align}
Z^{(i)}(t)&=Z^{(i)}(0)\exp\left[-(\lambda_1-\lambda_i)t\right]+\beta_{i1}\int_0^t\exp\left[ (\lambda_1-\lambda_i)(s-t)\right] dB(s).
\end{align}
%In this case, since $\lambda_1>\lambda_i, ~i=2,...,2d$, using the part result in analysis phase \Rmnum{1}, we know that each component is stable.
%And from the update we can see that even with $H_1(0)=h_0^1=0$, $h_1^1$ with high probability to be not zero, then it will still converge to the SDE.
%\vspace{-0.025in}
By the property of the O-U process, we characterize the expectation and variance of $Z^{(i)}(t)$ for $i\neq 1$.
\begin{align}
&\EE Z^{(i)}(t) =Z^{(i)}(0)\exp \left[ -(\lambda_1-\lambda_i)t \right],\notag \\
&\EE \left(Z^{(i)}(t)\right)^2 =\frac{\beta^2_{i1}}{2(\lambda_1-\lambda_i)}+\Bigg[\Big(Z^{(i)}(0)\Big)^2\notag-\frac{\beta^2_{i1}}{2(\lambda_1-\lambda_i)}\Bigg]\exp\left[-2(\lambda_1-\lambda_i)t\right].
\end{align}
%\vspace{-0.025in}
Recall that the distribution of $z^{(i)}_{\eta,k}$ can be well approximated by the normal distribution of $Z^{(i)}(t)$ for a sufficiently small step size. This further implies that after sufficiently many iterations, SGD enforces $z^{(i)}_{\eta,k}\rightarrow 0$ except $i=1$. Meanwhile, SGD behaves like a biased random walk towards the optimum, when it iterates within a small neighborhood the optimum. But unlike Phase I, the variance gradually becomes a constant.

Based on theorem \ref{Convergence}, we further establish an iteration complexity bound for SGD in following proposition. 
%\vspace{-0.025in}
\begin{proposition}\label{Time-Optimal}
Given a pre-specified $\epsilon>0$, a sufficiently small $\eta$, and $\delta$ defined in Proposition \ref{Time_Saddle}, after restarting counter of iteration, we need at most
\begin{align*}
N_3=\frac{\eta^{-1}}{2(\lambda_1-\lambda_2)}\log\left(\frac{4(\lambda_1-\lambda_2)\delta^2}{(\lambda_1-\lambda_2)\epsilon\eta^{-1}-4d\max\limits_{1\leq i \leq d}\beta_{i1}^2}\right),
\end{align*}
iterations such that $\sum_{i=2}^{2d} \left(h_{\eta,N_3}^{(i)}\right)^2\leq\epsilon$ with probability at least $3/4$.
\end{proposition}
%\vspace{-0.1in}
The proof of Proposition \ref{Time-Optimal} is provided in Appendix \ref{Time-Optimal-proof}. Combining Propositions \ref{Time_Saddle}, \ref{Time_Deterministic}, and \ref{Time-Optimal}, we obtain a more refined result in the following corollary.

%\vspace{-0.025in}
\begin{corollary}\label{final-rate}
Given a sufficiently small pre-specified $\epsilon>0$, we choose
\begin{align*}
\eta \asymp \frac{\epsilon (\lambda_1-\lambda_2)}{d\max_{1\leq i \leq d} \beta_{i1}^2}.
\end{align*}
We need at most
\begin{align*}
N=O\left[ \frac{d}{\epsilon (\lambda_1-\lambda_2)^2}\log\left(\frac{d}{\epsilon}\right) \right]
\end{align*}
iterations such that we have $\norm{u_{\eta,n}-\hat{u}}_2^2 + \norm{v_{\eta,n}-\hat{v}}_2^2\leq 3\epsilon$ with probability at least $\frac{3}{4}.$
\end{corollary}
%\vspace{-0.1in}

The proof of Corollary \ref{final-rate} is provided in Appendix \ref{final-rate-proof}. We can further improve the probability to $1-\nu$ for some $\nu>0$ by repeating $\cO(\log 1/\nu)$ replicates of SGD. We then compute the geometric median of all output solutions. See more details in \cite{cohen2016geometric}. %Note that if we further impose a bounded fourth order moment assumption, the optimization error in Corollary \ref{final-rate} can be dimension-free.

\subsection{Extension to $m\neq d$}\label{mnotequald}

Our analysis can further extend to the case where $X$ and $Y$ have different dimensions, i.e., $m\neq d$. Specifically, we consider an alternative way to construct $P$ defined in \eqref{Trans}. We follow the same notations to Assumption~\ref{Assumption2}, and use $O_X$ and $O_Y$ to denote the transition matrix between the observed data and latent variables. The dimensions of $O_X$ and $O_Y$, however, are different now, i.e., $O_X\in \RR^{m\times m}$ and $O_Y\in \RR^{d\times d}$. Without loss of generality, we assume $m>d$ and $O_X=(\tilde{O}_X ~O_X^0)$, where $\tilde{O}_X\in \RR^{m\times d}$ and $O_X^0\in \RR^{m\times(m-d)}$, and $O_Y$ are the transform matrix of $X$ and $Y$, respectively. Then we have the singular value decomposition as follows,
\begin{align}
O_X^\top \Sigma_{XY} O_Y=D, \quad \textrm{where}~D=\left(
\begin{array}{c}
\tilde{D}\\
0
\end{array}
\right)~\textrm{and}~\tilde{D}=\textrm{diag}(\lambda_1,\lambda_2,...,\lambda_d).
\end{align}
Thus, we have $\tilde{O}_X^\top \Sigma_{XY} O_Y= \tilde{D}$ and $(O_X^0)^\top \Sigma_{XY} O_Y=0$. Now we design the orthogonal transform matrix $P$.
\begin{align}\label{square-construction}
P=\left(
\begin{array}{ccc}
\frac{1}{\sqrt{2}}\tilde{O}_X& O_X^0 & \frac{1}{\sqrt{2}}\tilde{O}_X\\
\frac{1}{\sqrt{2}}O_Y&  0&  -\frac{1}{\sqrt{2}}O_Y
\end{array}
\right).
\end{align}
One can check that
\begin{align}
\left(
\begin{array}{cc}
0 & \Sigma_{XY}\\
\Sigma_{XY}^\top & 0
\end{array}
\right)=P 
 \left(
\begin{array}{cc}
D & 0 \\
0 & -D^\top
\end{array}
\right) P^\top=P
\left(
\begin{array}{ccc}
\tilde{D} & 0 & 0\\
0 & 0& 0\\
0 & 0& -\tilde{D}
\end{array}
\right)
P^\top. 
\end{align}
Then our previous analysis using ODE and SDE still holds.

Note that for $d=m$, any column vector of $P$ in \eqref{Trans} is a stationary solution. Here the square matrix $P$ in \eqref{square-construction} contains $m+d$ column vectors, but only the first $d$ and last $d$ column vectors are stationary solutions. This is because the remaining $m-d$ column vectors are even not feasible solutions, and violate the constraint $v^\top v=1$. Thus, given a feasible initial, the algorithm will not be trapped in the subspace spanned by th remaining $m-d$ column vectors .
%There is also another issue with $m\neq d$. The algorithm will get stuck, if the initial solution is in the null space of $\Sigma_{XY}$. There are two approaches to address this issue: (1) We can adopt a random initialization. Since the measurement of the null space is essentially zero, we get an initial solution, which is not in the null space with probability one; (2) We can inject a Gaussian noise to the stochastic gradient in each iteration. This additional perturbation guarantees the solution to escape from the null space with probability one.

\subsection{Extension to Missing Values}

Our methodology and theory can tolerate missing values. For simplicity, we assume the entries of $X$ and $Y$ misses independently with probability $1-p$ in each iteration, where $p\in(0,1)$. We then set all missing entries as $0$ values. We denote such imputed vectors by $\tilde{X}_k$ and $\tilde{Y}_k$. One can verify $\frac{1}{p^2}\tilde{X}_k\cdot \tilde{Y}^\top_k$ is an unbiased estimator of $\Sigma_{XY}=\EE X_k Y_k^\top$. Note that $1/p^2$ can be further absorbed into the step size $\eta$, denoted by $\eta_p$. Then \eqref{free-update} becomes:
\begin{align}
u_{k+1}=u_k+\eta_p\left(\tilde{X_k}\tilde{Y_k}^\top v_k -u_k^\top \tilde{X_k}\tilde{Y_k}^\top v_k  u_k \right) \quad \textrm{and}\quad v_{k+1}=v_k+\eta_p\left(\tilde{Y_k}\tilde{X_k}^\top u_k -u_k^\top \tilde{X_k}\tilde{Y_k}^\top v_k v_k \right).\label{mis-free-update}
\end{align}
The convergence analysis is very similar to the standard setting with a different choice of $\eta_p$, and therefore is omitted.%In the next section, we show that for the missing value, our algorithm still works.
%As we can see the only difference between the case of missing data and that of complete data is the coefficient before the $\Sigma_{XY}$. 

%Furthermore, the samples in the following iterations are also orthogonal to the null space, and therefore, with probability $1$, the algorithm

%Then we claim that using the update formula \eqref{SGR-update-u}+\eqref{SGR-update-v} or \eqref{PSG-update-u-v} as long as one iteration is not in the null space, with probability $1$ the following iterations are also not in the null space. The idea behind this is that null space and the rest space are perpendicular, and there update can be done independently, which means if there is some component in the non-zero space, with probability $1$, it will not get into the null space at all. 

%!TEX root = ./draft.tex
\section{Numerical Experiments}\label{Simulations}

We first provide a simple example to illustrate our theoretical analysis. Specifically, we choose $m = d = 3$. We first generate the joint covariance matrix for the latent factors $\overline{X}$ and $\overline{Y}$ as
\begin{align*}
\Cov(\overline{X}) = \Sigma_{\overline{X}\overline{X}} = \left[
\begin{array}{ccc}
6 &2 &1\\
2 &6 &2\\
1 &2 &6
\end{array}
\right],\quad
\Cov(\overline{X},\overline{Y})=\Sigma_{\overline{X}\overline{Y}} = \left[
\begin{array}{ccc}
4 &0 &0\\
0 &2 &0\\
0 &0 &0.5
\end{array}
\right],
\end{align*}
and $\Sigma_{\overline{Y}\overline{Y}} = \Sigma_{\overline{X}\overline{X}}$. We then generate two matrices $\tilde{U}$ and $\tilde{V}$ with each entry independently sampled from $N(0,1)$. Then we convert $\tilde{U}$ and $\tilde{V}$ to orthonormal matrices $U$ and $V$ by Grand-Schmidt transformation. At last, we generate the joint covariance matrix for the observational random vectors $X$ and $Y$ using the following covariance matrix
\begin{align*}
\Cov(X) = U^\top\Sigma_{\overline{X}\overline{X}}U,\quad\Cov(X,Y) = U^\top\Sigma_{\overline{X}\overline{Y}}V,\quad \textrm{and}\quad\Cov(Y) =V^\top\Sigma_{\overline{Y}\overline{Y}}V.
\end{align*}
We consider the total sample size as $n=2\times10^{5}$ and choose $\eta=5\times 10^{-5}$. The initialization solution $(u_{0},v_{0})$ is a pair of singular vectors associated with the second largest singular value of $\Sigma_{XY}$, i.e., saddle point. We repeat the simulation with update \eqref{free-update} for $100$ times, and plot the obtained results.

\begin{figure*}[htb!]

\centering
\subfigure[All Three Phases of $h_k^{(1)}$.]{
	\includegraphics[width=2.0in]{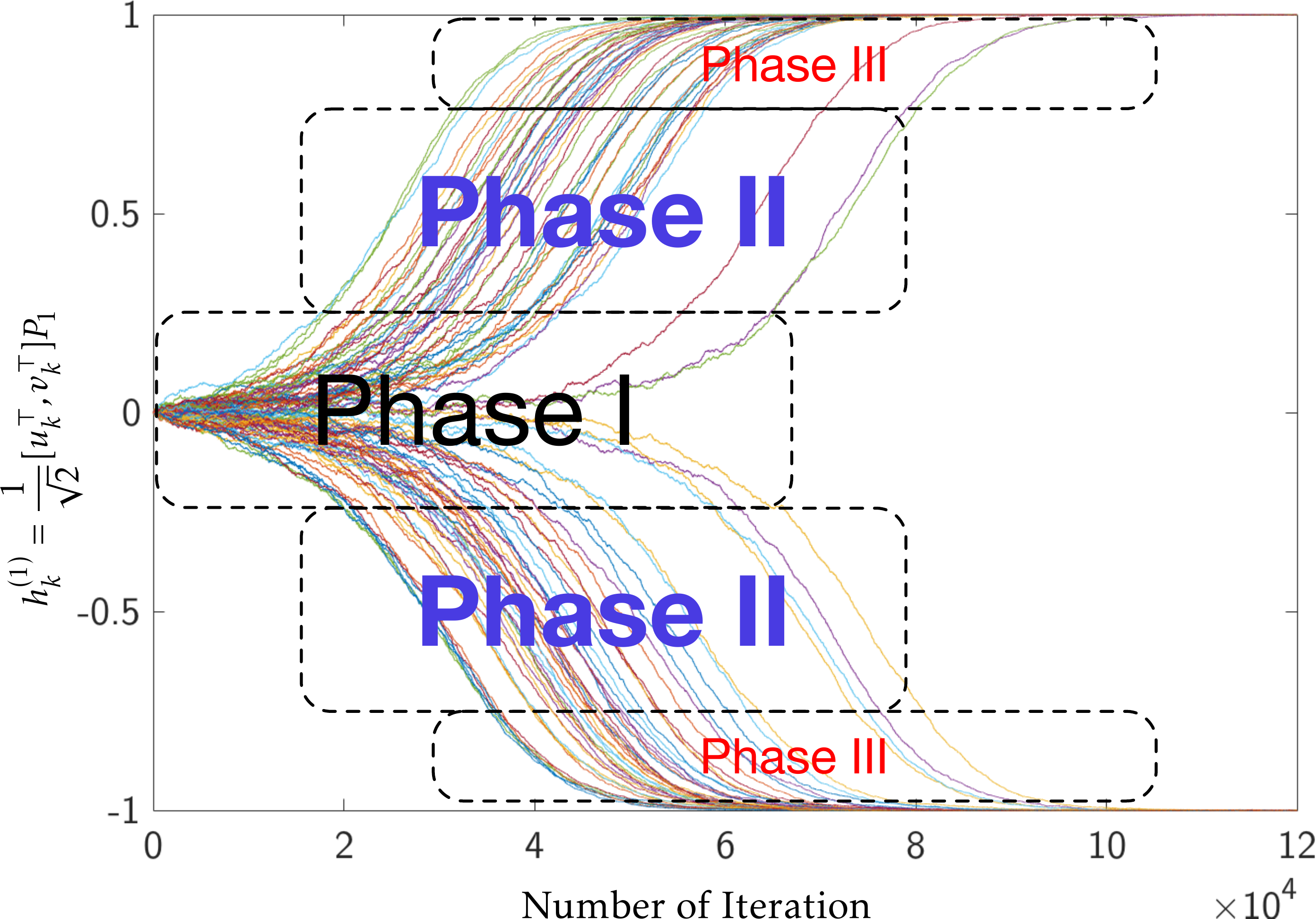}\label{all_phases_h1}
}
%\hspace{0.4in}
\subfigure[Phase I of $h_k^{(1)}$.]{
	\includegraphics[width=2.0in]{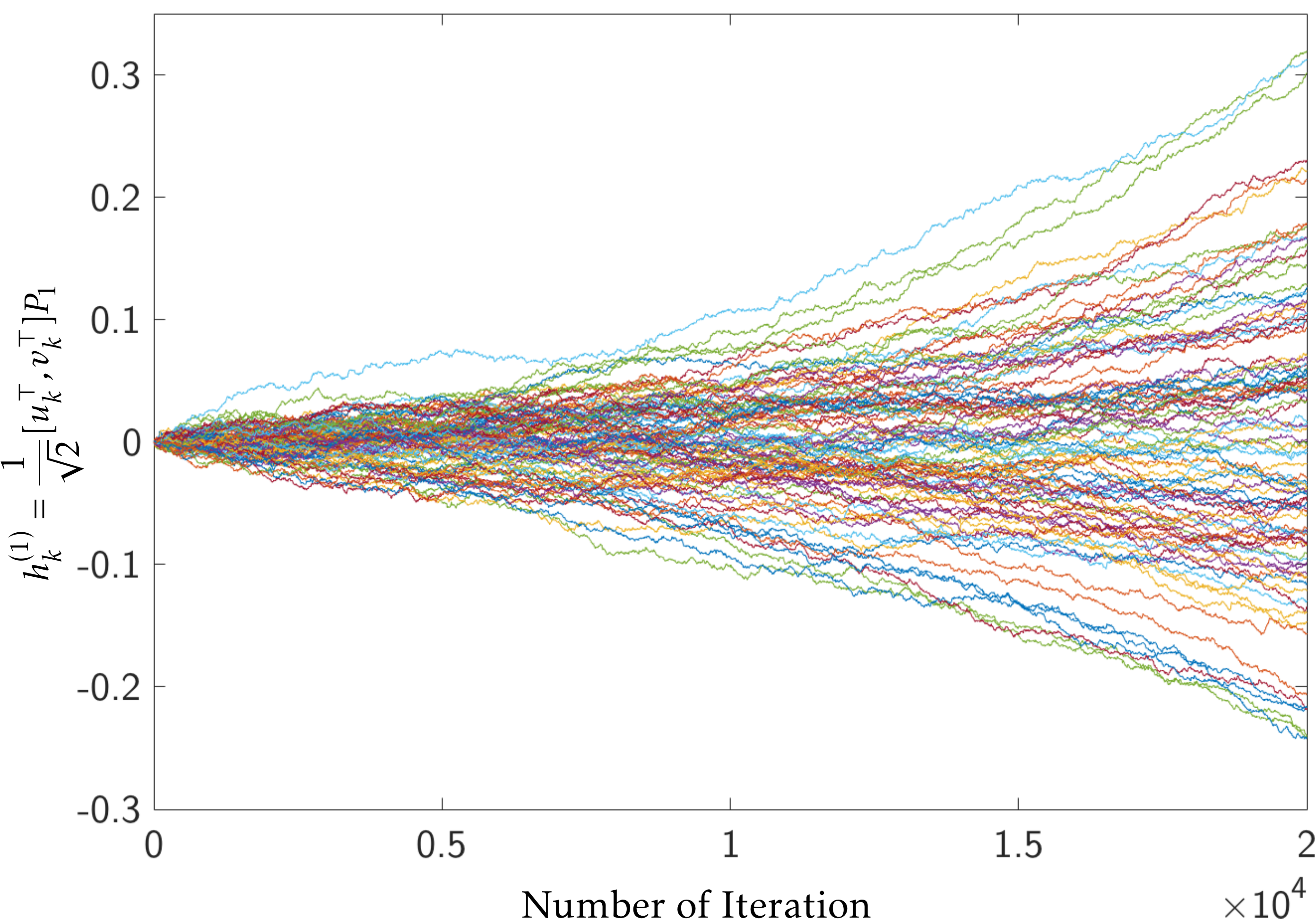}\label{phase_1_h1}
}
\subfigure[All Three Phases of $h_k^{(2)}$.]{
	\includegraphics[width=2.0in]{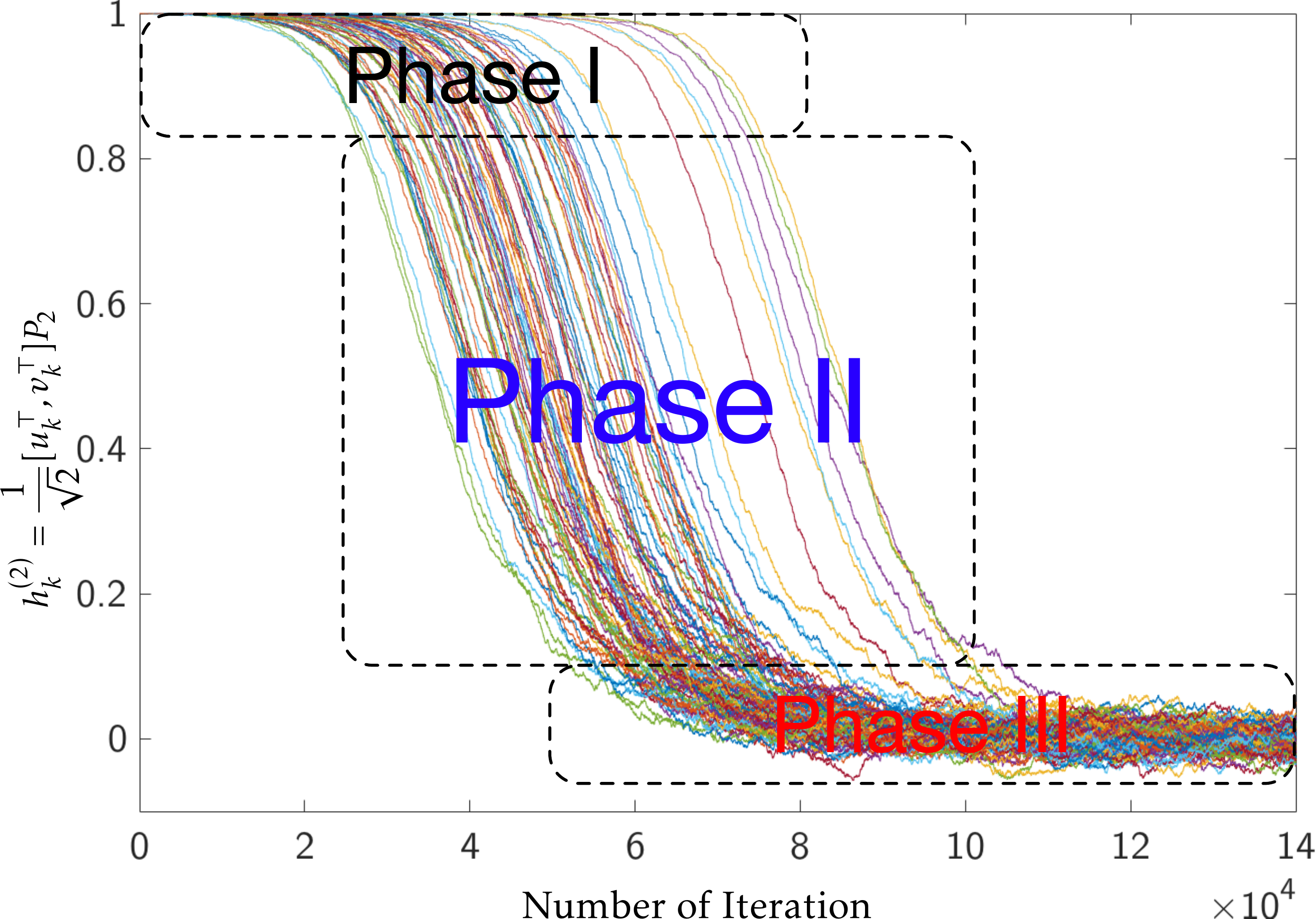}\label{all_phases_h2}
}

\caption{\it An illustrative example of the stochastic gradient algorithm. The three phases of the algorithm are consistent with our theory: In Phase I, the algorithm gradually escapes from the saddle point; In Phase II, the algorithm quickly iterates towards the optimum; In Phase III, the algorithm gradually converges to the optimum.}

\end{figure*}

Figure \ref{all_phases_h1} illustrates the three phases of the SGD algorithm. Specifically, the horizontal axis is the number of iterations, and the vertical axis is $h_k^{(1)}$ defined in \eqref{transformed_solution_w}. As $h_k^{(1)} \rightarrow \pm 1$, we have $u_k\rightarrow \pm\hat{u}$ and $v_k\rightarrow \pm\hat{v}$, e.g., global optima. This is due to the symmetric structure of the problem as mentioned in Section \ref{introduction}. Figure \ref{all_phases_h1} is consistent with our theory: In Phase I, the algorithm gradually escapes from the saddle point; In Phase II, the algorithm quickly moves towards the optimum; In Phase III, the algorithm gradually converges to the optimum.

Figure \ref{phase_1_h1} further zooms in Phase I of Figure \ref{all_phases_h1}. We see that the trajectories of all $100$ simulations behave very similar to an O-U process. Figure \ref{all_phases_h2} illustrates the three phases by $h_k^{(2)}$. As our analysis suggests, when $h_k^{(1)} \rightarrow \pm 1$, we have $h_k^{(2)} \rightarrow 0$. We see that the trajectories of all $100$ simulations also behave very similar to an O-U process in Phase III. These experimental results are consistent with our theory.

Also, we illustrate $h^{(1)}$ in Phase I and $h^{(2)}$ in Phase III are O-U processes by showing that 100 simulations of $h^{(1)}$ follow gaussian distributions at $10$-th, $100$-th, and $1000$-th iteration and those of $h^{(1)}$ follow gaussian distributions at $10^5$-th, $1.5\times 10^5$-th, and $2\times 10^5$-th iteration. This is consistent with the Theorems \ref{SDE for saddle} and \ref{Convergence} in Section \ref{SDE}. Also as we can see that in the Phase I, the variance of $h^{(1)}$ becomes larger and larger when the iteration number increases. Similarly, in the Phase III, the variance of $h^{(2)}$ becomes closer to a fixed number.

\begin{figure}[htb!]

\centering
\subfigure[The estimated density of $h^{(1)}$ in Phase I.]{
	\includegraphics[width=2.5in]{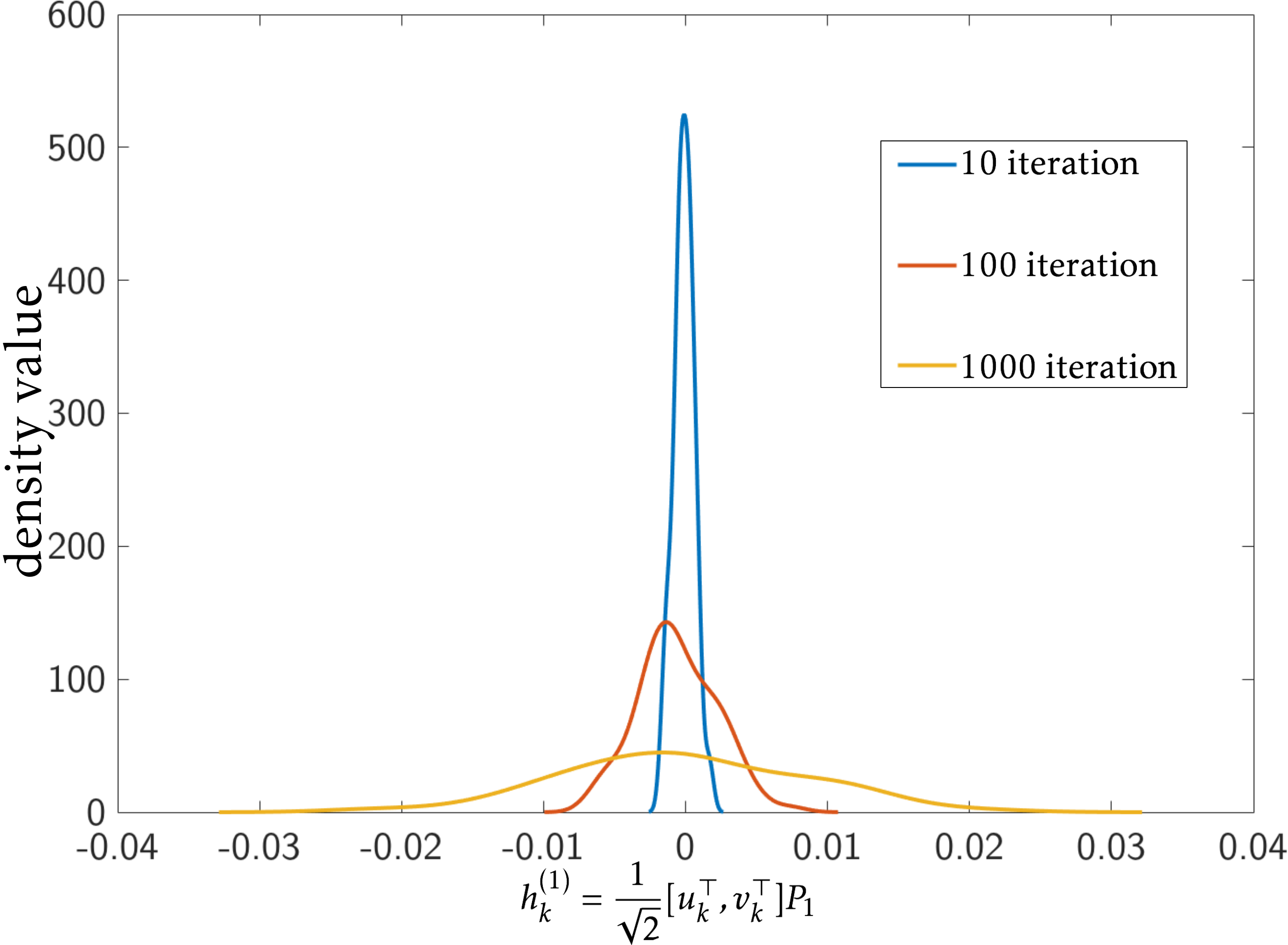}\label{h1}
}
\hspace{0.4in}
\subfigure[The estimated density of $h^{(2)}$ in Phase III.]{
	\includegraphics[width=2.5in]{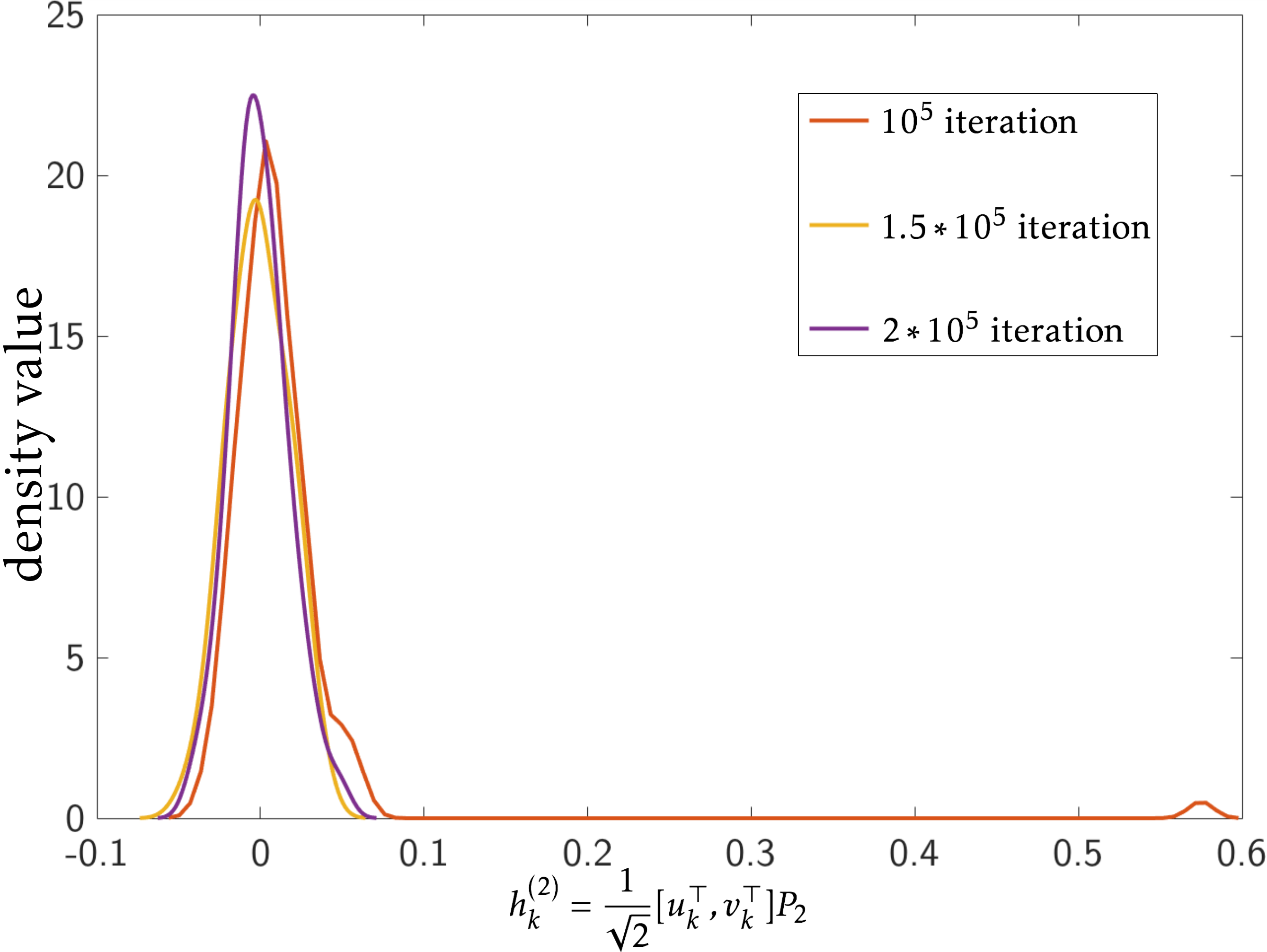}\label{h3}
}

\label{KDE}
\caption{\it The estimated density based on 100 simulations (obtained by kernel density estimation using 10-fold cross validation) at different iterations in Phase I and Phase III shows that $h_k^{(1)}$'s in Phase I and $h^{(2)}_k$'s in Phase III behave very similar to O-U processes. how their their variance change, which is consistent our theory.}
\end{figure}

\begin{figure}[htb!]

\centering
\subfigure[Comparison by Iteration.]{
	\includegraphics[width=2.5in]{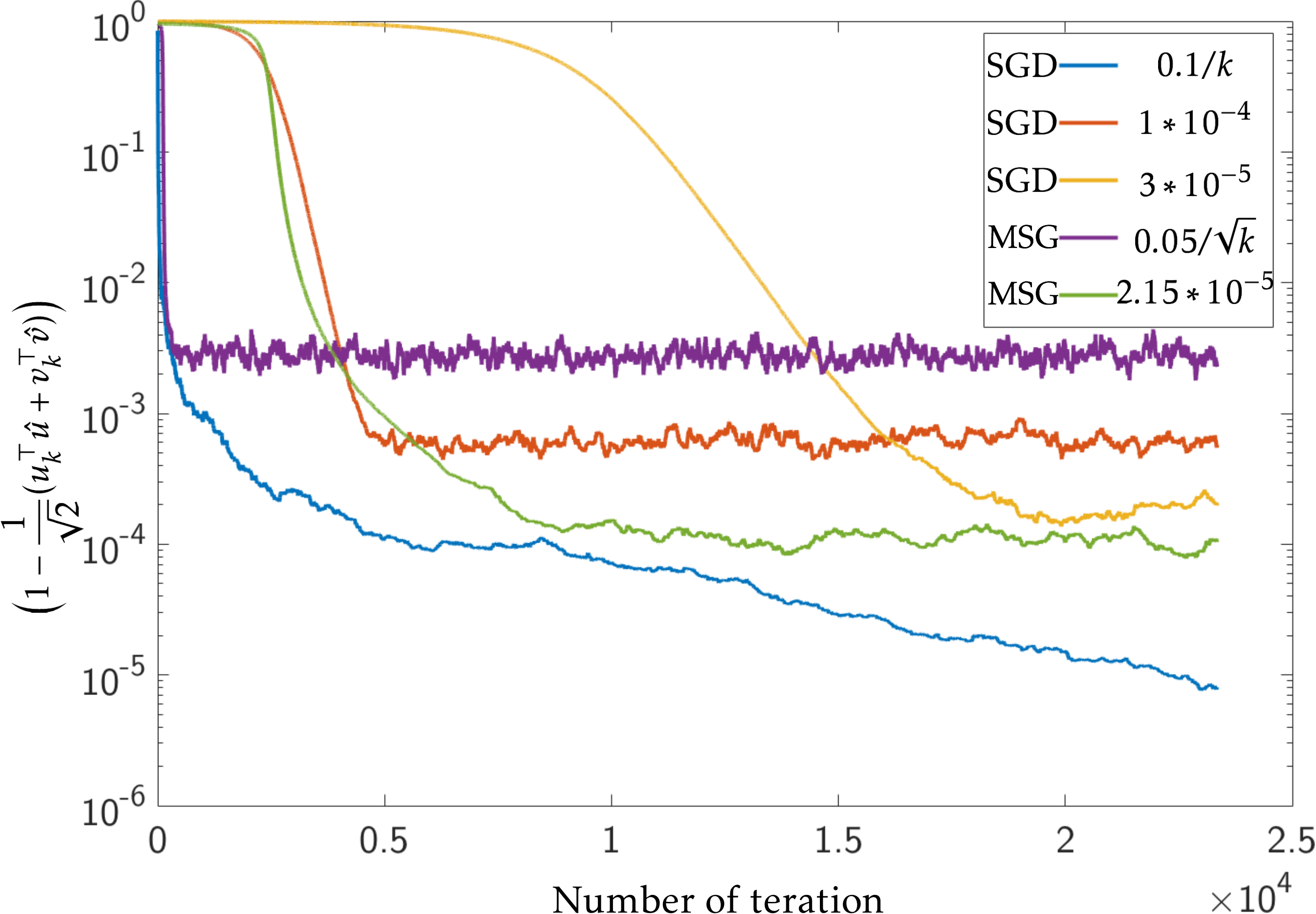}\label{iteration}
}
\hspace{0.4in}
\subfigure[Comparison by Time.]{
	\includegraphics[width=2.5in]{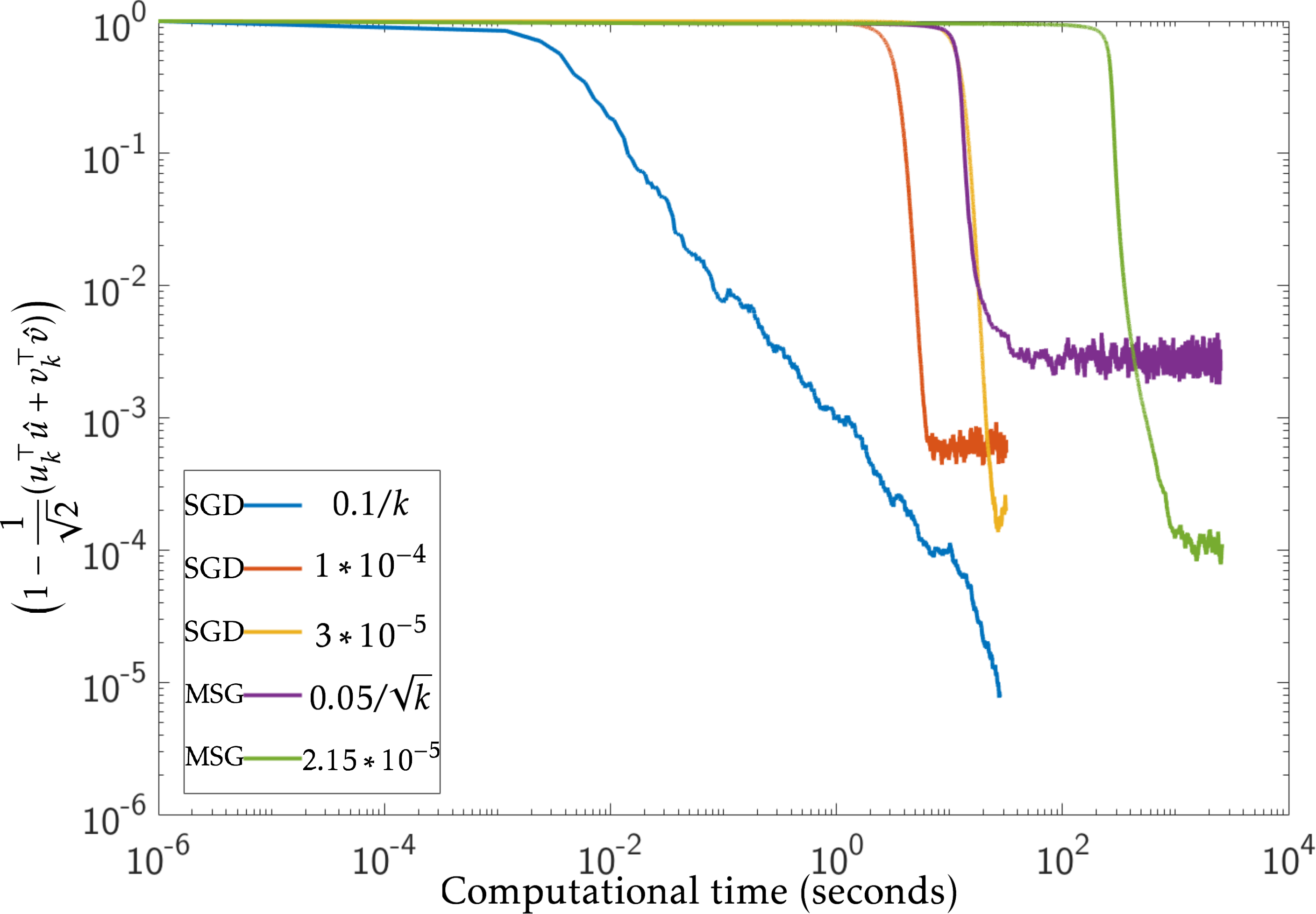}\label{time}
}
\caption{\it Comparison between nonconvex SGD and convex MSG with different step sizes. We see that SGD not only has a better iteration complexity, but also is more computationally efficient in wall clock time than convex MSG.}
\label{comparison}

\end{figure}

We then provide a real data experiment for comparing the computational performance our nonconvex stochastic gradient algorithm for solving \eqref{PLS-obj-1} with the convex stochastic gradient algorithm for solving \eqref{PLS-cvx}. We choose a subset of the MNIST dataset, whose labels are $3,4,5,~\textrm{or}~9$. The total sample size is $n=23343$, and $m=d=392$.  As \cite{arora2016stochastic} suggest, we choose $\eta_k = 0.05/\sqrt{k}$ or $2.15\times 10^{-5}$, for the convex stochastic gradient algorithm. For our nonconvex stochastic gradient algorithm, we choose either $\eta_k = 0.1/k$, $10^{-4}$, or $3\times10^{-5}$. Figure \ref{comparison} illustrates the computational performance in terms of iterations and wall clock time. As can be seen, our nonconvex stochastic gradient algorithm outperforms the convex counterpart in iteration complexity, and significantly outperforms in wall clock time, since the nonconvex algorithm does not need the computationally expensive projection in each iteration. This suggests that dropping convexity for PLS can boost both computational scalability and efficiency.

\begin{figure}

\centering
\subfigure[Different missing probabilities with step size $p^2*10^{-4}$.]{
	\includegraphics[width=2.5in]{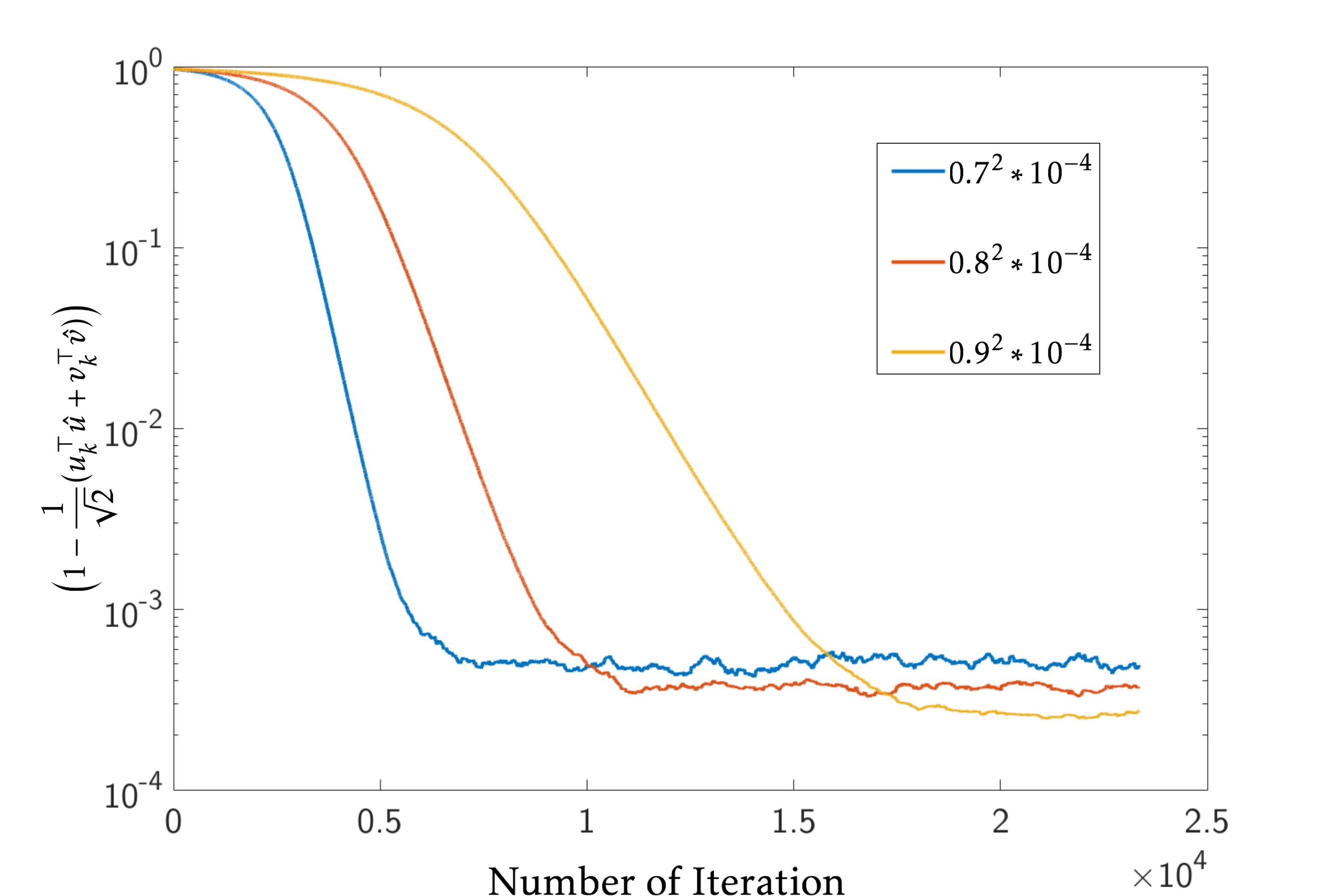}\label{prob}
}
\hspace{0.4in}
\subfigure[Different step sizes with missing probability $0.1$ (i.e., $p=0.9$).]{
	\includegraphics[width=2.5in]{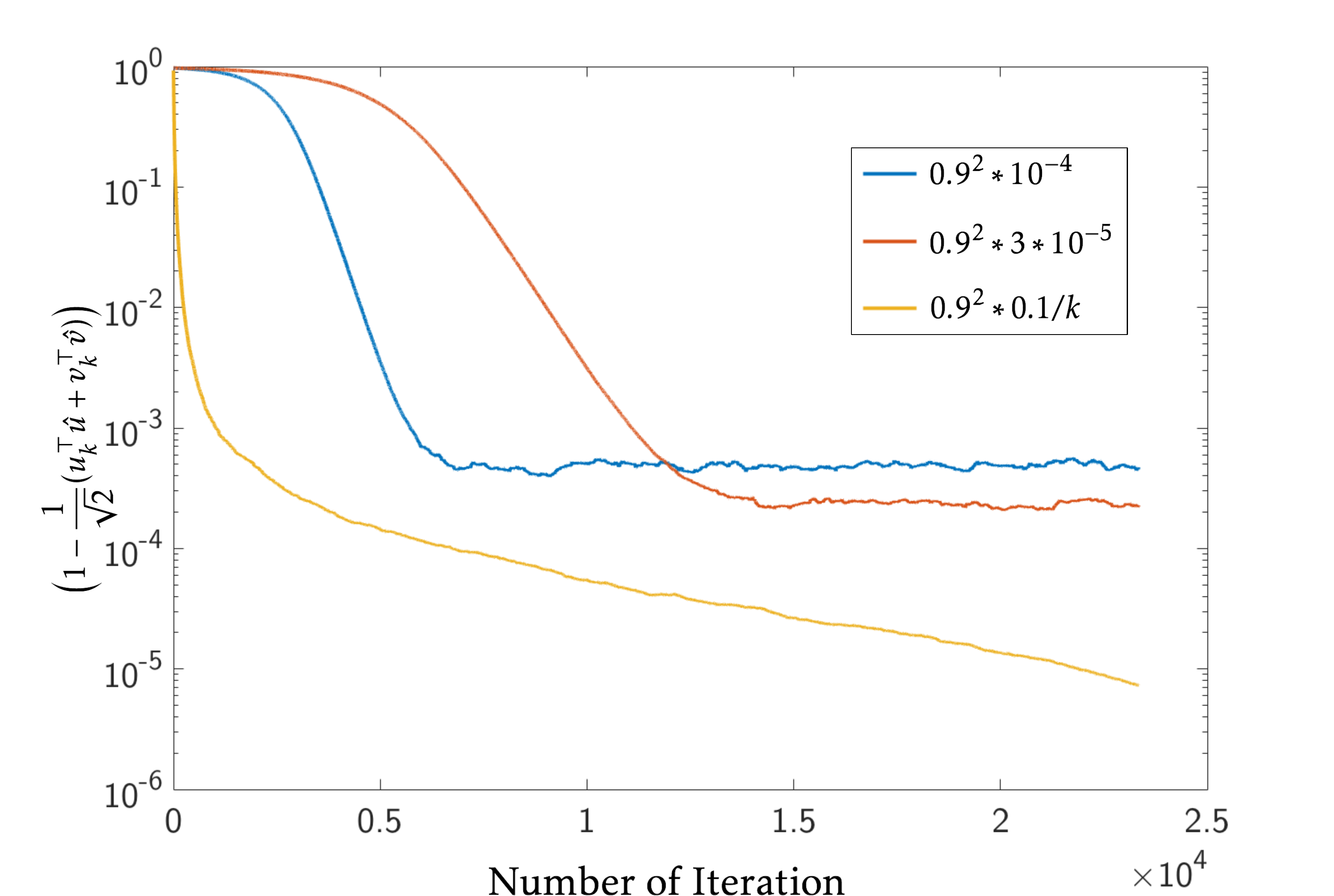}\label{stepsize}
}
\caption{\it Comparison among different missing probabilities and step sizes.}
\label{Missing}
\end{figure} 

Our last experiment demonstrates the computational performance of our proposed SGD algorithm when there exist missing values. Specifically, we adopt the same MNIST data set as our previous experiment. We independently drop each pixel of the image in each iteration with probability $(1-p)$. Figure \ref{Missing} illustrates the computational performance in terms of iterations under different missing probability and choices of the step size parameter. As can be seen, the empirical convergence of our proposed SGD algorithm is similar to (but slower than) that of our previous experiment without missing values.

\section{Discussions}\label{Discussions}

We establish the convergence rate of stochastic gradient descent (SGD) algorithms for solving online partial least square (PLS) problems based on diffusion process approximation. Our analysis indicates that for PLS, dropping convexity actually improves efficiency and scalability. Our convergence results are tighter than existing convex relaxation based method by a factor of $O(1/\epsilon)$, where $\epsilon$ is a pre-specified error. We believe the following directions should be of wide interests:
\begin{enumerate}
\item Our current results hold only for the top pair of left and right singular vectors, i.e., $r=1$. For $r>1$, we need to solve
\begin{align}
\label{eqn:def-r}
(\hat{U},~\hat{V}) = \argmax_{U\in \RR^{m\times r}, V\in\RR^{d\times r}}~\EE~\tr(V^\top YX^\top U)\quad\textrm{subject~to}\quad U^\top U =I_r,\quad V^\top V =I_r.
\end{align}
%where we rewrite $U$ as a vector $u\in\RR^m$ and $v\in\RR^d$. 
%We will explain more details on the rank-$r$ case in the later section.
Our approximations using ODE and SDE, however, do not admit unique solution due to rotation or permutation. Thus, extension of our analysis to $r>1$ is a challenging, but also an important future direction.

\item Our current results are only applicable to a fixed step size $\eta\asymp \epsilon (\lambda_1-\lambda_2)d^{-1}$. Our experiments suggest that the diminishing step size $\eta_k\asymp k^{-1}(\lambda_1-\lambda_2)^{-1}\log d$, $k$ from 1 to $N$, where $N$ is the sample complexity from theory, achieves a better empirical performance. One possible probability tool is Stein's method \citep{ross2011fundamentals}.

\item Our current results rely on the classical central limit theorem-type analysis by taking $\eta\rightarrow0^+$. Note the analysis of $\norm{u}_2=\norm{v}_2=1$ is an asymptotic result, and in experiment, when $\eta$ is small, $u$ and $v$ exactly stay on the sphere. But to get a more general result, connecting our analysis to discrete algorithmic proofs such as \cite{jain2016streaming, shamir2015fast,li2016near} should be an important direction \citep{barbour2005introduction}. One possible probability tool for addressing this issue is Stein's method \citep{ross2011fundamentals}.

%\item Our current results depend on the eigen-gap of problem. \cite{allen2016first} provide some results with an eigen-gap free bound.
\end{enumerate}
Moreover, our proposed SGD algorithm for PLS is also closely related to Canonical Correlation Analysis. Specifically, CCA solves a similar problem
\begin{align}\label{CCA-obj}
(\hat{u},\hat{v})=&\argmax_{u,v}~u^\top{\EE}XY^\top v\quad\text{subject to}\quad \EE(X^\top u)^2=1,~\EE(Y^\top v)^2=1.
\end{align}
For notational simplicity, we denote $\Sigma_{XY}={\EE}XY^\top$, $\Sigma_{XX}=\EE XX^\top$, and $\Sigma_{YY}=\EE YY^\top$. Since computing $\EE XX^\top$ and $\EE YY^\top$ is not affordable, the projected stochastic gradient algorithms are not applicable. Thus we consider an alternative approach to avoid the projection operation. We consider the Lagrangian function of \eqref{CCA-obj} as
\begin{align}\label{CCA-Lagrangian}
L(u,v,\mu,\sigma) = u^\top\Sigma_{XY} v &-\mu(u^\top\Sigma_{XX}u-1)-\sigma(v^\top\Sigma_{YY} v-1),
\end{align}
where $\mu$ and $\sigma$ are Lagrangian multipliers. We then check the optimal KKT conditions,
\begin{align*}
\Sigma_{XY} v - 2\Sigma_{XX}\mu u = 0,\quad\Sigma_{XY} u - 2\Sigma_{YY}\sigma v = 0,\quad u^\top\Sigma_{XX} u = 1\quad\textrm{and}\quad  v^\top\Sigma_{YY} v = 1,
\end{align*}
which further imply
\begin{align*}
u^\top\Sigma_{XY} v - 2\mu u^\top\Sigma_{XX} u = u^\top\Sigma_{XY} v-2\mu=0\quad\textrm{and}\quad v^\top\Sigma_{XY} u - 2\sigma v^\top\Sigma_{YY} v = v^\top{\EE}YX^\top u - 2\sigma= 0.
\end{align*}
Solving the above equations, we obtain the optimal Lagrangian multipliers as 
\begin{align}\label{CCA-optimal-multiplier}
\mu = \sigma = \frac{1}{2}u^\top\EE XY^\top v.
\end{align}
%Plugging \eqref{CCA-optimal-multiplier} into \eqref{CCA-Lagrangian}, we can convert the original optimization problem to
%\begin{align}\label{CCA-unconstrained}
%\max_{u,v}~2u^\top\Sigma_{XY} v &-\frac{1}{2}u^\top\Sigma_{XY} v\cdot u^\top\Sigma_{XX} u-\frac{1}{2}u^\top\Sigma_{XY} v\cdot v^\top\Sigma_{YY} v.
%\end{align}
Similarly, we then apply the dual free stochastic gradient method to solve \eqref{CCA-obj}. Specifically, at the $k$-th iteration, we independently sample $(X_k,Y_k)$ and $(\tilde{X}_k,\tilde{Y}_k)$ from $\cD$. Then we obtain
\begin{align}
u_{k+1} = u_k + \eta\Big(\tilde{X}_k\tilde{Y}_k^\top v_k -u_k^\top X_kY_k^\top v_k\cdot\tilde{X}_k\tilde{X}_k^\top u_k\Big),~~ v_{k+1} = v_k + \eta\Big( \tilde{Y}_k\tilde{X}_k^\top u_k -v_k^\top Y_kX_k^\top u_k\cdot \tilde{Y}_k\tilde{Y}_k^\top v_k\Big).\label{CCA-SGD}
\end{align}
Here we sample two pairs of $X$ and $Y$ to ensure the unbiasedness of the stochastic gradient. 

Then we can convert \eqref{CCA-SGD} to ordinary differential equations by taking $\eta\rightarrow0^+$, we get
\begin{align*}
\frac{dU}{dt} = \Sigma_{XY}  V -U^\top \Sigma_{XY}  V\cdot\Sigma_{XX} U,\quad\frac{dV}{dt}= \Sigma_{XY}^\top U -V^\top  \Sigma_{XY}^\top U\cdot  \Sigma_{YY}V.
\end{align*}
Different from PLS, the above ordinary differential equations do not admit a closed form solution, which makes our ODE/SDE-type convergence analysis not applicable in a straightforward manner. A possible alternative approach is to establish the lower bounds for $|\hat{u}^\top U(t)|$ and $|\hat{v}^\top V(t)|$, and further prove that as $t\rightarrow\infty$, we have $U(t)\rightarrow\hat{u}$ and $V(t)\rightarrow\hat{v}$. We will leave this option for further investigation.

Taking our result for PLS as an initial start, we expect more sophisticated and stronger follow-up work that applies to CCA and other online optimization problems with similar structures, which eventually benefits the learning community in both practice and theory.
%Existing literature has investigate the equivalence between CCA and SVD. For simplicity, we assume $\Sigma_{XX}$ and $\Sigma_{YY}$ are invertible. We define $w=\Sigma_{XX}^{1/2}u$, $z = \Sigma_{YY}^{1/2}v$, and $\Gamma =  \Sigma_{XX}^{-1/2}\Sigma_{XY}\Sigma_{YY}^{-1/2}$. Then \eqref{CCA-obj} can be rewritten as
%\begin{align}\label{CCA-obj-1}
%(\hat{w},\hat{z})=\argmax_{w,z}~w^\top\Gamma z\quad\text{subject to}\quad \norm{w} =1,~ \norm{z} =1.
%\end{align}
%Accordingly, we write \eqref{CCA-SGR-update-u} and \eqref{CCA-SGR-update-v} as
%\begin{align*}
%\end{align*}

\bibliography{ref}

\begin{thebibliography}{30}
\expandafter\ifx\csname natexlab\endcsname\relax\def\natexlab#1{#1}\fi
\expandafter\ifx\csname url\endcsname\relax
  \def\url#1{\texttt{#1}}\fi
\expandafter\ifx\csname urlprefix\endcsname\relax\def\urlprefix{URL }\fi

\bibitem[{Abdi(2003)}]{abdi2003partial}
\textsc{Abdi, H.} (2003).
\newblock Partial least square regression (pls regression).
\newblock \textit{Encyclopedia for research methods for the social sciences}
  792--795.

\bibitem[{Ando and Zhang(2005)}]{ando2005framework}
\textsc{Ando, R.~K.} and \textsc{Zhang, T.} (2005).
\newblock A framework for learning predictive structures from multiple tasks
  and unlabeled data.
\newblock \textit{Journal of Machine Learning Research} \textbf{6} 1817--1853.

\bibitem[{Arora et~al.(2012)Arora, Cotter, Livescu and
  Srebro}]{arora2012stochastic}
\textsc{Arora, R.}, \textsc{Cotter, A.}, \textsc{Livescu, K.} and
  \textsc{Srebro, N.} (2012).
\newblock Stochastic optimization for pca and pls.
\newblock In \textit{Communication, Control, and Computing (Allerton), 2012
  50th Annual Allerton Conference on}. IEEE.

\bibitem[{Arora and Livescu(2012)}]{arora2012kernel}
\textsc{Arora, R.} and \textsc{Livescu, K.} (2012).
\newblock Kernel cca for multi-view learning of acoustic features using
  articulatory measurements.
\newblock In \textit{MLSLP}. Citeseer.

\bibitem[{Arora et~al.(2016)Arora, Mianjy and Marinov}]{arora2016stochastic}
\textsc{Arora, R.}, \textsc{Mianjy, P.} and \textsc{Marinov, T.} (2016).
\newblock Stochastic optimization for multiview representation learning using
  partial least squares.
\newblock In \textit{Proceedings of The 33rd International Conference on
  Machine Learning}.

\bibitem[{Barbour and Chen(2005)}]{barbour2005introduction}
\textsc{Barbour, A.~D.} and \textsc{Chen, L. H.~Y.} (2005).
\newblock \textit{An introduction to Stein's method}, vol.~4.
\newblock World Scientific.

\bibitem[{Bharadwaj et~al.(2012)Bharadwaj, Arora, Livescu and
  Hasegawa-Johnson}]{bharadwaj2012multiview}
\textsc{Bharadwaj, S.}, \textsc{Arora, R.}, \textsc{Livescu, K.} and
  \textsc{Hasegawa-Johnson, M.} (2012).
\newblock Multiview acoustic feature learning using articulatory measurements.
\newblock In \textit{Intl. Workshop on Stat. Machine Learning for Speech
  Recognition}. Citeseer.

\bibitem[{Cai et~al.(2016)Cai, Li, Ma et~al.}]{cai2016optimal}
\textsc{Cai, T.~T.}, \textsc{Li, X.}, \textsc{Ma, Z.} \textsc{et~al.} (2016).
\newblock Optimal rates of convergence for noisy sparse phase retrieval via
  thresholded wirtinger flow.
\newblock \textit{The Annals of Statistics} \textbf{44} 2221--2251.

\bibitem[{Candes et~al.(2015)Candes, Li and Soltanolkotabi}]{candes2015phase}
\textsc{Candes, E.~J.}, \textsc{Li, X.} and \textsc{Soltanolkotabi, M.} (2015).
\newblock Phase retrieval via wirtinger flow: Theory and algorithms.
\newblock \textit{IEEE Transactions on Information Theory} \textbf{61}
  1985--2007.

\bibitem[{Chaudhuri et~al.(2009)Chaudhuri, Kakade, Livescu and
  Sridharan}]{chaudhuri2009multi}
\textsc{Chaudhuri, K.}, \textsc{Kakade, S.~M.}, \textsc{Livescu, K.} and
  \textsc{Sridharan, K.} (2009).
\newblock Multi-view clustering via canonical correlation analysis.
\newblock In \textit{Proceedings of the 26th annual international conference on
  machine learning}. ACM.

\bibitem[{Chen et~al.(2017)Chen, Yang, Li and Zhao}]{chen2017online}
\textsc{Chen, Z.}, \textsc{Yang, L.~F.}, \textsc{Li, C.~J.} and \textsc{Zhao,
  T.} (2017).
\newblock Online partial least square optimization: Dropping convexity for
  better efficiency and scalability.
\newblock In \textit{Proceedings of The 34th International Conference on
  Machine Learning}.

\bibitem[{Cohen et~al.(2016)Cohen, Lee, Miller, Pachocki and
  Sidford}]{cohen2016geometric}
\textsc{Cohen, M.~B.}, \textsc{Lee, Y.~T.}, \textsc{Miller, G.},
  \textsc{Pachocki, J.} and \textsc{Sidford, A.} (2016).
\newblock Geometric median in nearly linear time.
\newblock In \textit{Proceedings of the 48th Annual ACM SIGACT Symposium on
  Theory of Computing}. ACM.

\bibitem[{Dhillon et~al.(2011)Dhillon, Foster and Ungar}]{NIPS2011_4193}
\textsc{Dhillon, P.}, \textsc{Foster, D.~P.} and \textsc{Ungar, L.~H.} (2011).
\newblock Multi-view learning of word embeddings via cca.
\newblock In \textit{Advances in Neural Information Processing Systems 24}
  (J.~Shawe-Taylor, R.~S. Zemel, P.~L. Bartlett, F.~Pereira and K.~Q.
  Weinberger, eds.). Curran Associates, Inc., 199--207.

\bibitem[{Ethier and Kurtz(2009)}]{ethier2009markov}
\textsc{Ethier, S.~N.} and \textsc{Kurtz, T.~G.} (2009).
\newblock \textit{Markov processes: characterization and convergence}, vol.
  282.
\newblock John Wiley \& Sons.

\bibitem[{Evans(1988)}]{evans1988partial}
\textsc{Evans, W.} (1988).
\newblock Partial differential equations.

\bibitem[{Ge et~al.(2015)Ge, Huang, Jin and Yuan}]{ge2015escaping}
\textsc{Ge, R.}, \textsc{Huang, F.}, \textsc{Jin, C.} and \textsc{Yuan, Y.}
  (2015).
\newblock Escaping from saddle points-online stochastic gradient for tensor
  decomposition.
\newblock In \textit{COLT}.

\bibitem[{Golub and Van~Loan(2012)}]{golub2012matrix}
\textsc{Golub, G.~H.} and \textsc{Van~Loan, C.~F.} (2012).
\newblock \textit{Matrix computations}, vol.~3.
\newblock JHU Press.

\bibitem[{Hardoon et~al.(2004)Hardoon, Szedmak and
  Shawe-Taylor}]{hardoon2004canonical}
\textsc{Hardoon, D.~R.}, \textsc{Szedmak, S.} and \textsc{Shawe-Taylor, J.}
  (2004).
\newblock Canonical correlation analysis: An overview with application to
  learning methods.
\newblock \textit{Neural computation} \textbf{16} 2639--2664.

\bibitem[{Jain et~al.(2016)Jain, Jin, Kakade, Netrapalli and
  Sidford}]{jain2016streaming}
\textsc{Jain, P.}, \textsc{Jin, C.}, \textsc{Kakade, S.~M.},
  \textsc{Netrapalli, P.} and \textsc{Sidford, A.} (2016).
\newblock Streaming pca: Matching matrix bernstein and near-optimal finite
  sample guarantees for oja's algorithm.
\newblock In \textit{29th Annual Conference on Learning Theory}.

\bibitem[{Kidron et~al.(2005)Kidron, Schechner and Elad}]{kidron2005pixels}
\textsc{Kidron, E.}, \textsc{Schechner, Y.~Y.} and \textsc{Elad, M.} (2005).
\newblock Pixels that sound.
\newblock In \textit{Computer Vision and Pattern Recognition, 2005. CVPR 2005.
  IEEE Computer Society Conference on}, vol.~1. IEEE.

\bibitem[{Li et~al.(2016{\natexlab{a}})Li, Wang, Liu and Zhang}]{li2016near}
\textsc{Li, C.~J.}, \textsc{Wang, M.}, \textsc{Liu, H.} and \textsc{Zhang, T.}
  (2016{\natexlab{a}}).
\newblock Near-optimal stochastic approximation for online principal component
  estimation.
\newblock \textit{arXiv preprint arXiv:1603.05305} .

\bibitem[{Li et~al.(2016{\natexlab{b}})Li, Wang and Liu}]{li2016online}
\textsc{Li, C.~J.}, \textsc{Wang, Z.} and \textsc{Liu, H.}
  (2016{\natexlab{b}}).
\newblock Online ica: Understanding global dynamics of nonconvex optimization
  via diffusion processes.
\newblock In \textit{Advances in Neural Information Processing Systems}.

\bibitem[{Nowakowski(2013)}]{nowakowski2013multi}
\textsc{Nowakowski, B.~D.} (2013).
\newblock On multi-parameter semimartingales, their integrals and weak
  convergence .

\bibitem[{{\O}ksendal(2003)}]{oksendal2003stochastic}
\textsc{{\O}ksendal, B.} (2003).
\newblock Stochastic differential equations.
\newblock In \textit{Stochastic differential equations}. Springer, 65--84.

\bibitem[{Ross et~al.(2011)}]{ross2011fundamentals}
\textsc{Ross, N.} \textsc{et~al.} (2011).
\newblock Fundamentals of stein's method.
\newblock \textit{Probab. Surv} \textbf{8} 210--293.

\bibitem[{Sanger(1989)}]{sanger1989optimal}
\textsc{Sanger, T.~D.} (1989).
\newblock Optimal unsupervised learning in a single-layer linear feedforward
  neural network.
\newblock \textit{Neural networks} \textbf{2} 459--473.

\bibitem[{Shamir(2015)}]{shamir2015fast}
\textsc{Shamir, O.} (2015).
\newblock Fast stochastic algorithms for svd and pca: Convergence properties
  and convexity.
\newblock \textit{arXiv preprint arXiv:1507.08788} .

\bibitem[{Socher and Fei-Fei(2010)}]{socher2010connecting}
\textsc{Socher, R.} and \textsc{Fei-Fei, L.} (2010).
\newblock Connecting modalities: Semi-supervised segmentation and annotation of
  images using unaligned text corpora.
\newblock In \textit{Computer Vision and Pattern Recognition (CVPR), 2010 IEEE
  Conference on}. IEEE.

\bibitem[{Vinokourov et~al.(2002)Vinokourov, Shawe-Taylor and
  Cristianini}]{vinokourov2002inferring}
\textsc{Vinokourov, A.}, \textsc{Shawe-Taylor, J.} and \textsc{Cristianini, N.}
  (2002).
\newblock Inferring a semantic representation of text via cross-language
  correlation analysis.
\newblock In \textit{NIPS}, vol.~1.

\bibitem[{Zhao et~al.(2015)Zhao, Wang and Liu}]{zhao2015nonconvex}
\textsc{Zhao, T.}, \textsc{Wang, Z.} and \textsc{Liu, H.} (2015).
\newblock A nonconvex optimization framework for low rank matrix estimation.
\newblock In \textit{Advances in Neural Information Processing Systems}.

\end{thebibliography}
\bibliographystyle{ims}

\appendix

%!TEX root = ./draft.tex

%\onecolumn
%\appendix
\section{Proof Detailed Proofs in Section~\ref{CSS}}
\subsection{Proof of Proposition~\ref{geometry}}\label{geometry-proof}
\begin{proof}
We consider a compact singular value decomposition of $\Sigma_{XY}$ as follow:
\begin{align*}
\Sigma_{XY}=\sum_{i=1}^r \lambda_i \overline{u}_i \overline{v}_i^\top,
\end{align*}
where $\lambda_1>\lambda_2\geq...\geq\lambda_r>0$ are nonzero singular values, and $(\overline{u}_i,~\overline{v}_i)$'s are a pair of singular vectors associated with $\lambda_i$. Plugging \eqref{optimal-multiplier} into \eqref{KKT-Cond}, we have
\begin{align}\label{Station}
\Sigma_{XY}  v-(u^\top \Sigma_{XY}  v) u=0 \quad \textrm{and} \quad \Sigma_{XY} ^\top u- (u^\top \Sigma_{XY}  v) v=0.
\end{align}
%The next lemma shows that $(u,v)$ must be a pair of eigenvectors of $\Sigma_{XY}$, if 
%The next theorem characterizes the geometric properties of the stationary points.
%\begin{theorem}
%Suppose that $(u,v,\mu,\sigma)$ is a stationary point of \eqref{Lagrangian}. If $(u,v)$ is not a global optimum to \eqref{eqn:def}, then $(u,v)$ is an unstable stationary point, i.e.,
%\begin{align*}
%\lambda_{\max}(\nabla_{u,v}^2L(u,v,\mu,\sigma))>0.
%\end{align*}
%\end{theorem}
%By simple manipulation, we obtain
%\begin{align}
%\Sigma_{XY} ^\top \Sigma_{XY}  v-(u^\top \Sigma_{XY}  v) \Sigma_{XY} ^\top u &= \Sigma_{XY} ^\top \Sigma_{XY}  v-(u^\top \Sigma_{XY}  v)^2 v = 0\label{KKT-Cond1}\\
%\textrm{and}~\quad\Sigma_{XY} \Sigma_{XY} ^\top u- (v^\top \Sigma_{XY}  u) \Sigma_{XY} v&=\Sigma_{XY} \Sigma_{XY} ^\top u- (u^\top \Sigma_{XY}  v)^2 u=0\label{KKT-Cond2}.
%\end{align}
Since every vector $u\in \RR^m$ and $v\in\RR^d$ can be expanded as
\begin{align}\label{expand}
u=\sum_{i=1}^r c_i \overline{u}_i+\sum_{j=r+1}^m c_j \overline{u}_j \quad \textrm{and} \quad v=\sum_{i=1}^r l_i \overline{v}_i+\sum_{j=r+1}^d l_j \overline{v}_j,
\end{align}
where $\overline{u}_j$ for $j=r+1,...,m$ and $\overline{v}_j$ for $j=r+1,...,d$ are orthonormal basis vectors, and complementary to $\overline{u}_i$'s and $\overline{v}_i$'s for $i = 1,...,r$ in $\RR^m$ and $\RR^d$ respectively, and $c_i$'s and $l_i$'s are the coefficients. Plugging \eqref{expand} into the first equation of \eqref{Station}, we get
\begin{align}\label{Analy}
0&= \sum_{i=1}^r \lambda_i \overline{u}_i \overline{v}_i^\top \cdot \sum_{i=1}^d c_i\overline{v}_i-\sum_{i=1}^m l_i \overline{u}_i \cdot \sum_{i=1}^r \lambda_i \overline{u}_i \overline{v}_i^\top \cdot \sum_{i=1}^d c_i \overline{v}_i\cdot \sum_{i=1}^m l_i\overline{u}_i\notag\\
&= \sum_{i=1}^r c_i \lambda_i \overline{u}_i- \sum_{i=1}^m \big(\sum_{k=1}^r l_k \lambda_k c_k \big)\cdot l_i\overline{u}_i\notag\\
&= \sum_{i=1}^r \left(c_i \lambda_i -\big(\sum_{k=1}^r l_k \lambda_k c_k \big)\cdot l_i \right)\overline{u}_i-\sum_{i=r+1}^m \big(\sum_{k=1}^r l_k \lambda_k c_k \big)\cdot l_i\overline{u}_i.
\end{align}
The second equality holds because $\overline{u}_i$ and $\overline{v}_j$ are the columns of the orthogonal matrices. Since $\overline{u}_i$'s are the basis vectors of $\RR^m$, by \eqref{Analy}, we know the coefficients of all $\overline{u}_i$'s should be $0$. Therefore we consider two scenarios: 
\begin{enumerate}
\item If $\sum_{i=1}^r l_k \lambda_k c_k=0$, then we have $c_i=0$, $i=1,2,...,r$. Similarly, plugging \eqref{expand} into the second equation of \eqref{Station}, we have $l_i=0$, $i=1,2,...,r$. Thus, $u$ and $v$ are in the row and column null space of $\Sigma_{XY}$ respectively.
\item If $\sum_{i=1}^r l_k \lambda_k c_k\neq0$, then we have $l_i=0$, $i=r+1,...,m$, which further leads to:
\begin{align}\label{Deter}
c_i\lambda_i=\big(\sum_{k=1}^r l_k \lambda_k c_k \big)\cdot l_i \quad \textrm{and}\quad l_i\lambda_i=\big(\sum_{k=1}^r l_k \lambda_k c_k \big)\cdot c_i \quad \textrm{for $i=1,2,...,r$}.
\end{align}
Note that \eqref{Deter} holds if and only if there exists only one $i\in\{1,2,...,r\}.$ $c_j=l_j=\pm \delta_{ij},~j=1,2,...,r,$ where $\delta_{ij}$ is the Kronecker delta, i.e., $\delta_{ij}=\left\{
\begin{array}{cc}
1&\quad i=j\\
0& \quad i\neq j
\end{array}
\right..$
\end{enumerate}
The verification of the above points satisfying \eqref{Station} is straightforward, and therefore omitted.
%There are two implications: if $(u,v)$ satisfies \eqref{KKT-Cond1} and \eqref{KKT-Cond2}, then the following results hold:
%\begin{enumerate}
%\item $(u,v)$ must be the eigenvectors of $\Sigma_{XY} \Sigma_{XY} ^\top$ and $\Sigma_{XY}^\top \Sigma_{XY}$ respectively;
%\item $(u,v)$ must be associated with the same eigenvalue, i.e.,
%\begin{align*}
%u^\top\Sigma_{XY} \Sigma_{XY} ^\top u = v^\top\Sigma_{XY}^\top \Sigma_{XY}v = (u^\top \Sigma_{XY}  v)^2.
%%\end{align*}
%\end{enumerate}
%We find all the stationary points of \eqref{Lagrangian}.
\end{proof}
\subsection{Proof of Proposition~\ref{stable}}\label{stable-proof}
\begin{proof}
For notation simplicity, we denote $\nabla_{u,v}^2 L(u,v)$ as $\nabla_{u,v}^2 L(u,v,\mu,\sigma)\Big|_{\mu=\sigma=\frac{1}{2}u^\top A v}$
\begin{align*}
\nabla_{u,v}^2 L(u,v)=\left(
%\begin{array}{cc}
%-2\mu \cdot I_m & A \\
%A^\top & -2\sigma \cdot I_d
%\end{array}
%\right)=\left(
\begin{array}{cc}
-u^\top \Sigma_{XY} v \cdot I_m & \Sigma_{XY} \\
\Sigma_{XY}^\top & -u^\top \Sigma_{XY} v \cdot I_d
\end{array}
\right).
\end{align*}

\begin{enumerate}
\item[\bf{a.}] If $u$ and $v$ are in the row and column null space of $\Sigma_{XY}$ respectively, then
\begin{align*}
\nabla_{u,v}^2 L(u,v)=\left(
\begin{array}{cc}
0 & \Sigma_{XY} \\
\Sigma_{XY}^\top & 0 
\end{array}
\right)\quad\textrm{and}\quad\lambda_{\max}(\nabla_{u,v}^2 L(u,v)) = \lambda_1.
\end{align*}
Therefore, it is an unstable stationary point because of the positive curvature.

\item[\bf{b.}] If $(u,v)$ is a pair of singular vector of $\lambda_i$, then by simple linear algebra, we know that
\begin{align*}
\nabla_{u,v}^2 L(u,v) \sim \left(
\begin{array}{cc}
-u^\top \Sigma_{XY} v \cdot I_m & 0 \\
0 & \frac{1}{u^\top \Sigma_{XY} v}\Sigma_{XY}^\top \Sigma_{XY}-u^\top \Sigma_{XY} v \cdot I_d 
\end{array}
\right).
\end{align*}
%According to the KKT conditions, we have 
%\begin{align*}
%A^\top A v-(u^\top A v)^2 v=0 \quad \textrm{and} \quad AA^\top u- (u^\top A v)^2 u=0,
%\end{align*}
%which means at the stationary point, $u$ and $v$ are the eigen-vectors of $AA^\top$ and $A^\top A$ respectively corresponding to the same eigen-value $(u^\top A v)^2$. Without loss of generality, we assume $d\leq m$. There exist two orthogonal matrices $O_1$ and $O_2$ such that $O_1^\top A O_2=\left(
%\begin{array}{c}
%G \\
%0
%\end{array}\right),$ where $G=\diag(l_1,l_2,\dots,l_d)$, and $l_1,l_2,\dots,l_d$ are part of singular values. To illustrate the difference of these two type stationary points clearly, we assume $l_1>l_2>\cdots>l_d$. Then we have 
%\begin{align*}
%O_2^\top A^\top A O_2=G^2=\diag(l_1^2,\cdots,l_d^2) \quad\textrm{and}\quad O_1^\top AA^\top O_1=\left(
%\begin{array}{cc}
%G^2 & 0\\
%0 & 0 
%\end{array}\right).
%\end{align*}
One can verify %By $\Sigma_{XY}^\top \Sigma_{XY}\sim \diag(\lambda_1^2,...,\lambda_r^2,0,...,0)$, we further have
\begin{align*}
\lambda_{\max}(\nabla_{u,v}^2 L(u,v)) = \frac{\lambda_1^2-\lambda_i^2}{\lambda_i}\geq\lambda_1-\lambda_2.
\end{align*}
%\begin{align}
%\nabla_{u,v}^2 L(u,v)\sim \frac{1}{u^\top \Sigma_{XY} v}\left(
%\begin{array}{cc}
%-\lambda_i^2 \cdot I_m  & 0 \\
%0 &  \diag(\lambda_1^2-\lambda_i^2,...,\lambda_r^2-\lambda_i^2,-\lambda_i^2,...,-\lambda_i^2)
%\end{array}
%\right).
%\end{align}
Therefore, the Hessian matrix is negative semi-definite if and only if $u^\top \Sigma_{XY} v=\lambda_1$, i.e., $(u,v)$ is the optimum of \eqref{eqn:def}. The Hessian has a positive eigenvalue. 
%\begin{enumerate}
%\item[1)] $u^\top \Sigma_{XY} v=\lambda_i$.
%\item[2)] $u^\top \Sigma_{XY} v=-\lambda_i$.
%\end{enumerate}
\end{enumerate}
Thus, only the optima of \eqref{Lagrangian} are stable stationary points. All the others are unstable.
\end{proof}

\section{Proof Detailed Proofs in Section~\ref{ODE}}
\subsection{Proof of Theorem~\ref{Converge-ODE}}\label{Converge-ODE-proof}

\begin{proof}

First, we calculate the infinitesimal conditional expectation. Since the optimization problem is symmetric about $u$ and $v$, we only prove the claim for $u$,
\begin{align*}
\frac{d}{dt}\EE \left(U_\eta (t)-U_\eta(0)\right)\big|_{t=0}&=\eta^{-1} \EE\left(U_\eta (\eta)-U_\eta (0)\big|U_\eta (0),V_\eta (0) \right)\\
&=\Sigma_{XY} V(0)-U(0)^\top \Sigma_{XY} V(0) U(0).
\end{align*}
Next, we show that if the initial is on the sphere, then with probability $1$, all iterations are on the sphere as $\eta\rightarrow 0^+$. Given $\norm{u_k}_2=\norm{v_k}_2=1$, we have
\begin{align*}
\norm{u_{k+1}}_2^2 & =\left(u_k+\eta\cdot(X_kY_k^\top v_k -u_k^\top X_k Y_k^\top v_k u_k)\right)^\top\cdot \left(u_k+\eta\cdot(X_kY_k^\top v_k -u_k^\top X_k Y_k^\top v_k u_k)\right)\\
& = u_k^\top u_k +2\eta( u_k^\top X_k Y_k^\top v_k -u_k^\top X_k Y_k^\top v_k u_k^\top u_k)+\eta^2\norm{X_kY_k^\top v_k -u_k^\top X_k Y_k^\top v_k u_k}_2^2\\
& = 1+\eta^2\norm{X_kY_k^\top v_k -u_k^\top X_k Y_k^\top v_k u_k}_2^2.
\end{align*}
Therefore, we get
\begin{align*}
\PP\Big(\lim_{\eta\rightarrow0^+} \norm{u_{k+1}}_2=1 \Big| \norm{u_k}_2=1\Big)=\PP(|X_k^\top Y_k| <\infty)=1.
\end{align*}
The last equality holds, since $\EE|X_k^\top Y_k|$ is finite:
\begin{align*}
\EE|X_k^\top Y_k| \leq \sqrt{\EE\norm{X_k}_2^2\cdot\EE\norm{Y_k}_2^2}\leq B^2d.
\end{align*}
%(Since both $\EE\norm{X_k}_2^2$ and $\EE\norm{Y_k}_2^2$ are bound,  $\EE|X_k^\top Y_k|$ is bounded).
Finally, we bound the infinitesimal conditional variance.
\begin{align*}
& ~~~~~\frac{d}{dt}\EE \left(U_\eta^{(j)}(t)-U_\eta^{(j)}(0)\right)^2\big|_{t=0}\\
%& = \eta^{-1}\mathbb{E} \left[\left(U_\eta^{(j)}(\eta)-U_\eta^{(j)}(0)\right)^2 \Big | U_\eta(0)=u_k,~V_\eta(0)=v_k \right]\\
%& = \eta^{-1}\mathbb{E} \left[\left(U_\eta^{(j)}(\eta)-U_\eta^{(j)}(0)\right) \left(U_\eta^{(j)}(\eta)-U_\eta^{(j)}(0)\right) \Big | U_\eta(0)=u_k,~V_\eta(0)=v_k \right]  \\
& \leq \eta^{-1} \cdot \tr \left(\mathbb{E} \left[\left(U_\eta(\eta)-U_\eta(0)\right) \left(U_\eta(\eta)-U_\eta(0)\right)^\top\right)\Big | U_\eta(0)=u_k,~V_\eta(0)=v_k\right]\\
%& = \eta^{-1}\mathbb{E}\left[\left(U_\eta(\eta)-U_\eta(0)\right)^\top \left(U_\eta(\eta)-U_\eta(0)\right)\Big | U_\eta(0)=u_k,~V_\eta(0)=v_k\right]  \\
& = \eta^{-1}\cdot \EE \left[ \eta \left(X_kY_k^\top u_k-u_k^\top X_kY_k^\top v_k u_k\right)^\top\cdot\eta\left(X_kY_k^\top u_k-u_k^\top X_kY_k^\top v_k u_k\right)\right]\\
& =\eta \cdot \EE \left( u_k^\top Y_k X_k^\top X_k Y_k^\top u_k-2u_k^\top Y_k X_k^\top u_k u_k^\top X_kY_k^\top v_k + u_k^\top u_k (u_k^\top X_kY_k^\top v_k)^2  \right)\\
& \leq \eta\cdot \left(\sqrt{ \EE \norm{X_k}_2^4 \EE\norm{Y_k}_2^4 } +2\sqrt{ \EE(u_k^\top Y_k X_k^\top u_k )^2 \EE(u_k^\top Y_k X_k^\top v_k )^2} +\EE (u_k^\top X_kY_k^\top v_k)^2 \right)\\
& \leq \eta \cdot \left(\sqrt{ \EE \norm{X_k}_2^4 \EE\norm{Y_k}_2^4 } +3 \EE (|Y_k^\top| |X_k|) ^2  \right)\\
& = O(\eta).
\end{align*}
Last equality holds by the Assumption~\ref{Assumption1}.

%Furthermore, by Cauchy-Schwarz inequality, we have
%\begin{align*}
%\frac{d}{dt}\EE \left(U_\eta^{(j)}(t)-U_\eta^{(j)}(0)\right)^2\big|_{t=0} \leq \eta \EE \left((dB)^2+2(dB)^2+(dB)^2\right)+O(\eta^2)\leq \eta 4(dB)^2+O(\eta^2)= O(\eta).
%\end{align*}
Therefore, by Section 4 of Chapter 7 in \cite{ethier2009markov}, we know that, as $\eta \rightarrow 0^+$, $U_\eta(t)$ and $V_\eta(t)$ weakly converge to the solution of \eqref{limitation u-v} %and \eqref{limitation v} 
with the same initial. By definition of $U_\eta(t)$ and $V_\eta(t)$, we complete the proof.
\end{proof}

\subsection{Proof of Theorem~\ref{Solution-ODE}}\label{Solution-ODE-proof}

\begin{proof}
Since $P$ is an orthonormal matrix, $\norm{H_j}_2=\norm{W_j}_2=1$ for all $j=1,...,d$. Thus, we have
\begin{align*}
\frac{d}{dt}H^{(i)} &=\lambda_i H^{(i)}-\sum_{j=1}^{2d}\lambda_j(H^{(j)})^2 H^{(i)}  \notag \\
& = \lambda_i\sum_{j=1}^{2d}(H^{(j)})^2H^{(i)}-\sum_{j=1}^{2d}\lambda_j(H^{(j)})^2 H^{(i)} \notag \\
& = H^{(i)}\sum_{j=1}^{2d}\left(\lambda_i-\lambda_j\right)(H^{(j)})^2.
\end{align*}
We then verify \eqref{solution} satisfies \eqref{Simplified2}. By \cite{evans1988partial}, we know that since $H_j(t)$ is continuously differentiable in $t$, the solution to the ODE is unique. For notational simplicity, we denote 
\begin{align*}
S^{(j)}(t)=H^{(j)}(0)\exp(\lambda_jt).
\end{align*} Then we have
\begin{align*}
H^{(i)}(t)=\frac{S^{(i)}(t)}{\sqrt{\sum_{j=1}^{2d}\left(S^{(j)}(t)\right)^2}}.
\end{align*}
Now we only need to verify
\begin{align*}
\frac{d}{dt}H^{(i)}(t) & =\frac{\left(\lambda_iS^{(i)}(t)\right)\sqrt{\sum_{j=1}^{2d}\left(S^{(j)}(t)\right)^2}-\frac{\left(2\sum_{j=1}^{2d}\lambda_j\left(S^{(j)}(t)\right)^2\right)S^{(i)}(t)}{2\sqrt{\sum_{j=1}^{2d}\left(S^{(j)}(t)\right)^2}}}{\sum_{j=1}^{2d}\left(S^{(j)}(t)\right)^2}\notag \\
& = \lambda_i\frac{S^{(i)}(t)}{\sqrt{\sum_{j=1}^{2d}\left(S^{(j)}(t)\right)^2}}-\sum_{j=1}^{2d}\lambda_j\frac{\left(S^{(j)}(t)\right)^2}{\sum_{j=1}^{2d}\left(S^{(j)}(t)\right)^2}\frac{S^{(i)}(t)}{\sqrt{\sum_{j=1}^{2d}\left(S^{(j)}(t)\right)^2}}  \notag\\
& = \lambda_i H^{(i)}(t)-\sum_{j=1}^{2d}\lambda_j\left(H^{(j)}(t)\right)^2 H^{(i)}(t),
\end{align*}
which completes the proof.
\end{proof}

\section{Proof Detailed Proofs in Section~\ref{SDE}}

\subsection{Proof of Theorem~\ref{SDE for saddle}}\label{SDE for saddle-proof}
\begin{proof} We prove this by contradiction. Assume the conclusion does not hold, that is there exists a constant $C>0,$ such that for any  $\eta'>0$ we have $$\sup_{\eta\leq \eta'}\PP(\sup_t |Z_{\eta}^{(i)}(t)|\leq C)=1.$$ That implies there exists a sequence $\{\eta_n\}_{n=1}^\infty$ converging to $0$ such that 
\begin{align}\label{eq_contra}
\lim_{n\rightarrow\infty}\PP(\sup_t |Z_{\eta_n}^{(i)}(t)|\leq C)= 1.
\end{align}
Thus, condition (i) in Theorem 2.4 \citep{nowakowski2013multi} holds. We next check the second condition. When $\sup_t |Z_{\eta_n}^{(i)}(t)|\leq C$ holds, Assumption \ref{Assumption1} yields that $z_{\eta_n,k+1}^{(i)}-z_{\eta_n, k}^{(i)}=C'\eta_n,$ where $C'$ is some constant. Thus, for any $t,\epsilon>0,$ we have $$|Z^{(i)}_{\eta_n}(t)-Z^{(i)}_{\eta_n}(t+\epsilon)|=\frac{\epsilon}{\eta} C'\eta=C'\epsilon.$$ Thus, condition (ii) in Theorem Theorem 2.4 \citep{nowakowski2013multi} holds. Then we have  $\{Z^{(i)}_{\eta_n}(\cdot)\}_n$ is tight and thus converges weakly.

We then calculate the infinitesimal conditional expectation and variance for $Z_{\eta_n}^{(i)},~i\neq j$.
\begin{align}\label{Exp-Stage1}
\frac{d}{dt} \EE Z_{\eta_n}^{(i)} (t) \big|_{t=0}& =\eta_n^{-1}\EE \left[Z_{\eta_n}^{(i)}(\eta_n)-Z_{\eta_n}^{(i)} (0) \big|  H_{\eta_n} (0)=h\right]  \notag \\
& = \eta_n^{-1}\EE\left[\eta_n^{-1/2}\left(H_{\eta_n}^{(i)}(\eta_n)-H_{\eta_n}^{(i)} (0)\right) \big|  H_{\eta_n} (0) =h \right]\notag \\
&= \eta_n^{-1/2}h^{(i)}\sum_{l=1}^{2d}\left(\lambda_i-\lambda_l\right)(h^{(l)})^2 = Z_{\eta_n}^{(i)}\left(\lambda_i-\lambda_j\right)+o(1), %\qquad\quad\textrm{(since $ h \approx e_j, i\neq j$)}
\end{align}
where the last equality comes from the assumption that the algorithm starts near $j^{th}$ column of $P,~j\neq 1$, i.e., $h \approx e_j$. To compute variance, we first compute $\hat{\Lambda}$,
\begin{align*}
\hat{\Lambda} = P^\top Q P %= \frac{1}{\sqrt{2}}\left(
%\begin{array}{cc}
%O_X^\top & O_Y^\top\\
%O_X^\top &-O_Y^\top
%\end{array}
%\right) 
%\left(
%\begin{array}{cc}
%0  & XY^\top\\
%YX^\top & 0
%\end{array}
%\right) 
%\frac{1}{\sqrt{2}}\left(
%\begin{array}{cc}
%O_X & O_X\\
%O_Y &-O_Y
%\end{array}
%\right)
 = \frac{1}{2}\left(
\begin{array}{cc}
\overline{Y}~\overline{X}^\top+\overline{X}~\overline{Y}^\top & \overline{Y}~\overline{X}^\top- \overline{X}~\overline{Y}^\top \\
-\overline{Y}~\overline{X}^\top+\overline{X}~\overline{Y}^\top & -\overline{Y}~\overline{X}^\top-\overline{X}~\overline{Y}^\top
\end{array}
\right),
\end{align*}
where $Q$ is defined in \eqref{To-Simplify}.
Then we analyze $e_i^\top \hat{\Lambda} e_j$ by cases:
\begin{align*}
e_i^\top \hat{\Lambda} e_j & =\left\{
\begin{array}{ll}
\frac{1}{2} \left(\overline{X}^{(i)} \overline{Y}^{(j)}+ \overline{X}^{(j)} \overline{Y}^{(i)}\right) &\textrm{if}~\max(i,j)\leq d,\\
\frac{1}{2} \left(-  \overline{X}^{(j)} \overline{Y}^{(i-d)}+\overline{X}^{(i-d)} \overline{Y}^{(j)}\right) &\textrm{if}~  j\leq d<i,\\
\frac{1}{2} \left( \overline{X}^{(j-d)} \overline{Y}^{(i)}-\overline{X}^{(i)} \overline{Y}^{(j-d)}\right) &\textrm{if}~ i\leq d<j,\\
\frac{1}{2} \left(-\overline{X}^{(i-d)} \overline{Y}^{(j-d)}- \overline{X}^{(j-d)} \overline{Y}^{(i-d)}\right) &\textrm{if}~ \min(i,j)> d,
\end{array}
\right.% \notag \\
%& = \frac{1}{2}\left( -sgn(i-1/2-d)\overline{X}^{(j)}\overline{Y}^{(i)}-sgn(j-1/2-d)\overline{X}^{(i)}\overline{Y}^{(j)}\right)
\end{align*}
which further implies
\begin{align}\label{Var-Stage1}
\frac{d}{dt}\EE (Z_{\eta_n}^{(i)} (t)-Z_{\eta_n}^{(i)} (0))^2 \big|_{t=0}& =\eta_n^{-1}\EE \big[\big(Z_{\eta_n}^{(i)} (\eta)-Z_{\eta_n}^{(i)} (0)\big)^2 \big| H_{\eta_n}(0)=h\big] \notag \\
& = \eta_n^{-2}\EE [\eta_n^2(\hat{\Lambda}h-h^\top\hat{\Lambda}h h)(\hat{\Lambda}h-h^\top\hat{\Lambda}h h)^\top]_{i,i} \notag \\
& = \EE (e_i^\top\hat{\Lambda}e_j e_j^\top\hat{\Lambda}^\top e_i)+o(1) \notag  \\
%& = \left\{
%\begin{array}{lr}
%\frac{1}{4} \left(\gamma_i\omega_j+\gamma_j\omega_i+2\alpha_{i,j}\right), & sgn(i-d-1/2)*sgn(j-1/2-d)=1 \notag \\
%\frac{1}{4} \left(\gamma_i\omega_j+\gamma_j\omega_i-2\alpha_{i,j}\right), & sgn(i-d-1/2)*sgn(j-1/2-d)=-1 \notag
%\end{array}
%\right. \notag 
& = \frac{1}{4} \big(\gamma_i\omega_j+\gamma_j\omega_i+2\sign(i-d-1/2)\cdot \sign(j-1/2-d) \cdot \alpha_{ij}\big).
\end{align}
By \eqref{Exp-Stage1} and \eqref{Var-Stage1}, we get the limit stochastic differential equation,
\begin{equation*}
dZ^{(i)}(t)=-(\lambda_j-\lambda_i)Z^{(i)}(t)dt+\beta_{ij}dB(t).
\end{equation*}

Therefore, $\{Z^{(i)}_{\eta_n}(\cdot)\}$ converges weakly to a solution of	
% \begin{align}\label{SDE_1}
%  dU^i=\frac{\lambda_i-\lambda_j}{1-\mu}U^i dt+\frac{\alpha_{i,j}}{1-\mu}dB_t.
%  \end{align} 
  The process defined by the equation above is an unstable O-U process with mean $0$ and exploding variance. Thus, for any $\tau$, there exist a time $t'$, such that
$$\PP(|Z^{(i)}(t')|\geq C)\geq 2\tau.$$
Since $\{Z^{(i)}_{\eta_n}\}_n$ converges weakly to $Z^i,$ thus $\{Z^{(i)}_{\eta_n}(t')\}_n$ converges in distribution to $Z^{i}(t').$ This implies that there exists an $N>0$, such that for any $n>N$ 
$$|{\PP(|Z^i(T)|\geq C)-\PP(|Z^{(i)}_{\eta_n}(T)|\geq C)}|\leq {\tau}.$$
Then we find a $t'$ such that 
$$\PP(|Z^{(i)}_{\eta_n}(t')|\geq C)\geq {\tau}, \forall n>N, $$ 
or equivalently $$\PP(|Z^{(i)}_{\eta_n}(t')|\leq C)< 1-{\tau},  \forall n>N.$$ 
Since $\left\{\omega\big|\sup_t |Z_{\eta_n}^{(i)}(t)(\omega)|\leq C\right\}\subset\left\{\omega\big||Z^{(i)}_{\eta_n}(\tau')(\omega)<C\right\},$ we have 
$$\PP(\sup_t |H^{\eta_n,i}(t)|\leq C\sqrt{\eta_n})=\PP(\sup_t |Z_{\eta_n}^{(i)}(t)|\leq C)\leq 1-{\delta}, \forall n>N, $$ 
which leads to a contradiction with \ref{eq_contra}. Our assumption does not hold.

\end{proof}

%\begin{proof}%[Proof of Theorem~\ref{SDE for saddle}]
%\end{proof}

\subsection{Proof of Proposition~\ref{Time_Saddle}}\label{Time_Saddle-proof}

\begin{proof}%[Proof of Proposition \ref{Time_Saddle}]

Our analysis is based on approximating $z_{\eta,k}^{(1)}$ by its continuous approximation $Z_{\eta}^{(1)}(t)$, which is normal distributed at time $t$. By simple manipulation, we have
\begin{align*}
\PP\left((h_{\eta,N_1}^{(2)})^2\leq 1-\delta^2\right)  & = \PP\left((z_{\eta,N_1}^{(2)})^2\leq \eta^{-1}(1-\delta^2)\right) \geq \PP(|z_{\eta,N_1}^{(1)}| \geq \eta^{-\frac{1}{2}}\delta).%& = \PP\left((Z^{(2)}(\eta N_1))^2 \leq \eta^{-1}(1-\delta^2)\right)  \notag \\
\end{align*} 
We then prove $P\left(\left|z_{\eta,N_1}^{(1)}\right|\geq \eta^{-\frac{1}{2}}\delta\right)\geq 1-\nu$. At time t, $z_{\eta,k}^{(1)}$ approximates to a normal distribution with mean $0$ and variance $\frac{\beta_{12}^2}{2(\lambda_1-\lambda_2)}\left[\exp\big(2(\lambda_1-\lambda_2)\eta N_1\big)-1\right]$. Therefore, let $\Phi(x)$ be the CDF of $N(0,1)$, we have
\begin{align*}
\PP\left(\frac{\big |z_{\eta,N_1}^{(1)}\big |}{\sqrt{\frac{\beta_{12}^2}{2(\lambda_1-\lambda_2)} \cdot \left[\exp\left(2(\lambda_1-\lambda_2)\eta N_1\right)-1\right]}}\geq \Phi^{-1}\left(\frac{1+\nu}{2}\right)\right) \approx 1-\nu,
\end{align*}
which requires
$$\eta^{-\frac{1}{2}}\delta\leq \Phi^{-1}\left(\frac{1+\nu}{2}\right)\cdot \sqrt{\frac{\beta_{12}^2}{2(\lambda_1-\lambda_2)} \cdot \left[\exp\left(2(\lambda_1-\lambda_2) \eta N_1\right)-1\right]}.$$ Solving the above inequality, we get
%\begin{align*}
%&\eta^{-1}\delta^2  \leq \left(\Phi^{-1}\left(\frac{1+\nu}{2}\right)\right)^2 \cdot \frac{\beta_{1,2}^2}{2(\lambda_1-\lambda_2)} \cdot \left[\exp\left(2(\lambda_1-\lambda_2) \eta N_1\right)-1\right].
%\end{align*}
%After we simplify this inequality, we get $N_1$, that is 
\begin{align*}
N_1 = \frac{\eta^{-1}}{2(\lambda_1-\lambda_2)}\log\left(\frac{2\eta^{-1}\delta^2(\lambda_1-\lambda_2)}{\Phi^{-1}\left(\frac{1+\nu}{2}\right)^2\beta^2_{12}} +1\right).
\end{align*}
\end{proof}

\subsection{Proof of Proposition~\ref{Time_Deterministic}}\label{Time_Deterministic-proof}

\begin{proof}%[Proof of Proposition \ref{Time_Deterministic}]
After Phase I, we restart our counter, i.e., $h_{\eta,0}^{(1)}=\delta$. By \eqref{solution} and $h_{\eta,N_2}^{(1)}$ approximating to the process $H^{(1)}(\eta N_2)$, we obtain
\begin{align*}
\left(h_{\eta,N_2}^{(1)}(t)\right)^2 & = \left(H^{(1)}(\eta N_2)\right)^2 =\left( \sum\limits_{j=1}^{2d}\left(\left(H^{(j)}(0)\right)^2\exp{(2\lambda_j \eta N_2)}\right)\right)^{-1}\left(H^{(1)}(0)\right)^2\exp{(2\lambda_1 \eta N_2)} \notag \\
& \geq  \left( \delta^2 \exp(2\lambda_1 \eta N_2)+(1-\delta^2)\exp(2\lambda_2 \eta N_2) \right)^{-1}\delta^2 \exp(2\lambda_1 \eta N_2),
\end{align*}
which requires $$ \left( \delta^2 \exp(2\lambda_1 \eta N_2)+(1-\delta^2)\exp(2\lambda_2 \eta N_2) \right)^{-1}\delta^2 \exp(2\lambda_1\eta N_2) \geq \eta^{-1}(1-\delta^2).$$ Solving the above inequality, we get
\begin{align*}
N_2=\frac{\eta^{-1}}{2(\lambda_1-\lambda_2)}\log\frac{1-\delta^2}{\delta^2}~.
\end{align*}
\end{proof}

\subsection{Proof of Theorem~\ref{Convergence}}\label{Convergence-proof}
\begin{proof}%[Proof of Theorem~\ref{Convergence}] 
For $i=2,...,2d$, we compute the infinitesimal conditional expectation and variance,
\begin{align*}%\label{Exp-Stage3}
\frac{d}{dt} \EE Z_{\eta_n}^{(i)} (t) \big|_{t=t_0}& =\eta^{-1}\EE \left[Z_{\eta_n}^{(i)}(t_0+\eta)-Z_{\eta_n}^{(i)} (t_0)\big| H^\eta (t_0)=h\right] \notag \\
&= \eta^{-1/2}h_i\sum_{j=1}^{2d}\left(\lambda_i-\lambda_j\right)h_j^2 + O(\eta)= Z^{(i)}\left(\lambda_i-\lambda_1\right)+o(1),\notag\\
\frac{d}{dt}\EE \left(Z_{\eta_n}^{(i)} (t)-Z_{\eta_n}^{(i)} (t_0)\right)^2 \big|_{t=t_0} & =\eta^{-1}\EE \left[\left(Z_{\eta_n}^{(i)} (t_0+\eta)-Z_{\eta_n}^{(i)} (t_0)\right)^2 \big| H^\eta(t_0)=h\right] \notag \\
& = \eta^{-2}\EE \left[\eta^2(\hat{\Lambda}h-h^\top\hat{\Lambda}h h)(\hat{\Lambda}h-h^\top\hat{\Lambda}h h)^\top\right]_{i,i}+O(\eta) \notag \\
& = \EE (e_i^\top\hat{\Lambda}e_1 e_1^\top\hat{\Lambda}^\top e_i)+o(1) =\frac{1}{4} \left(\gamma_i\omega_1+\gamma_1\omega_i-2\sign(i-d-1/2) \alpha_{i1}\right)+o(1).
%& = \left\{
%\begin{array}{cc}
% \EE (\overline{X}^{(i)}\overline{Y}^{(1)}\overline{Y}^{(1)}\overline{X}^{(i)}), \quad i \leq d\\
% 0, \qquad ~~~~~~~~~~~~~~~~~~~~~~~~~~i>d
% \end{array}
% \right.  \notag\\
%& = \left\{
%\begin{array}{cc}
% Var(\overline{X}^{(i)}) Var(\overline{Y}^{(1)}), \quad i \leq d\\
% 0,  \qquad ~~~~~~~~~~~~~~~~~~~~~~~~~~i>d
% \end{array}
% \right. \notag \\
%& = \left\{
%\begin{array}{cc}
%\gamma_i \omega_j, \quad i \leq d \\
% 0,  \qquad~~i>d.
% \end{array}
% \right.
\end{align*}
Following similar lines to the proof of Theorem~\ref{SDE for saddle}, by Section 4 of Chapter 7 in \cite{ethier2009markov}, we have for each $k = 2,...,2d,$ if $Z^{(i)}(0)=\eta^{-1/2}h_{\eta,0}^{(i)}$ as $\eta \rightarrow 0^+$, then the stochastic process $\eta^{-1/2}h_{\eta,\lfloor t\eta^{-1}\rfloor}^{(k)}$ weakly converges to the solution of the stochastic differential equation~\eqref{Converge-Solution}.
\end{proof}

\subsection{Proof of Proposition~\ref{Time-Optimal}}\label{Time-Optimal-proof}

\begin{proof}

Since we restart our counter, we have $\sum_{i=2}^{2d} (z_{\eta,0}^{(i)})^2=\eta^{-1} \delta^2$. Since $z_{\eta,k}^{(i)}$ approximates to $Z^{(i)}(\eta k)$ and its second moment:
\begin{align*}
\EE \left(Z^{(i)}(t)\right)^2=\frac{\beta^2_{i1}}{2(\lambda_1-\lambda_i)}+\left(\left(Z^{(i)}(0)\right)^2-\frac{\beta^2_{i1}}{2(\lambda_1-\lambda_i)}\right)\exp\left[-2(\lambda_1-\lambda_i)t\right],\quad \textrm{for } i\neq 1,
\end{align*}
we use the Markov inequality:
\begin{align*}
\PP\left(\sum_{i=2}^{2d} \left(h_{\eta,N_3}^{(i)}\right)^2> \epsilon \right) &  \leq \frac{\EE \left(\sum_{i=2}^{2d} \left(h_{\eta,N_3}^{(i)}\right)^2 \right) }{\epsilon} = \frac{\EE \left(\sum_{i=2}^{2d} \left(z_{\eta,N_3}^{(i)}\right)^2 \right) }{\eta^{-1}\epsilon}  \notag \\
& =\frac{1}{\eta^{-1} \epsilon} \sum_{i=2}^{2d}\frac{\beta^2_{i1}}{2(\lambda_1-\lambda_i)}\Big(1-\exp\big(-2(\lambda_1-\lambda_i) \eta N_3\big)\Big)+\left(z_{\eta,0}^{(i)}\right)^2\exp\left[-2(\lambda_1-\lambda_i)\eta N_3\right]\notag\\
& \leq \frac{1}{\eta^{-1}\epsilon} \Big(\frac{d \max\limits_{2\leq i\leq d}(\beta_{i1}^2)}{2(\lambda_1-\lambda_2)}\Big(1-\exp\big(-2(\lambda_1-\lambda_d) \eta N_3\big)\Big) \notag \\
&~~~~~~~~~~~~~~~+\frac{d \max\limits_{d+1\leq i\leq 2d}(\beta_{i1}^2)}{2(\lambda_1+ \lambda_d)}\Big(1-\exp\big(-4\lambda_1 \eta N_3\big)\Big)+\delta^2\exp\left[-2(\lambda_1-\lambda_2)\eta N_3\right] \Big) \notag\\
& \leq  \frac{1}{\eta^{-1}\epsilon}\left(\frac{d \max\limits_{1\leq i\leq d}(\beta_{i1}^2)}{(\lambda_1- \lambda_2)}+\delta^2\exp\left[-2(\lambda_1-\lambda_2)\eta N_3\right] \right).
\end{align*}
To guarantee $\frac{1}{\eta^{-1} \epsilon}\left(\frac{d \max\limits_{1\leq i\leq d}(\beta_{i1}^2)}{(\lambda_1- \lambda_2)}+\delta^2\exp\left[-2(\lambda_1-\lambda_2)\eta N_3\right] \right) \leq \frac{1}{4}$, we get:
\begin{align*}
N_3 \geq \frac{\eta^{-1}}{2(\lambda_1-\lambda_2)}\log\left(\frac{4(\lambda_1-\lambda_2)\delta^2}{(\lambda_1-\lambda_2)\epsilon\eta^{-1}-4d\max\limits_{1\leq i \leq d}\beta_{i1}^2}\right).
\end{align*}

\end{proof}

\subsection{Proof of Corollary~\ref{final-rate}}\label{final-rate-proof}
\begin{proof}
%Since we already know that $\norm{u_{\eta,n}-\hat{u}}_2^2+\norm{v_{\eta,n}-\hat{v}}_2^2=2\norm{h_{\eta,n}-\hat{h}}_2^2$.
First, we prove that $\norm{u_{\eta,k}-\hat{u}}_2^2 + \norm{v_{\eta,k}-\hat{v}}_2^2$ can be bounded by $3\sum_{i=2}^{2d}\left(h_{\eta,k}^{(i)}\right)^2$, when it is near the optima. Recall that $h_{\eta,k}=\frac{1}{\sqrt{2}}P^\top (u_{\eta,k}^\top~ v_{\eta,k}^\top)^\top$ and $e_1=\hat{h}=\frac{1}{\sqrt{2}}\PP(\hat{u}^\top~ \hat{v}^\top)^\top$. Our analysis has shown that when $k$ is large enough, the SGD iterates near the optima. Then we have
\begin{align}\label{error}
\norm{u_{\eta,k}-\hat{u}}_2^2 + & \norm{v_{\eta,k}-\hat{v}}_2^2  = 4-2\langle u_{\eta,k},\hat{u}\rangle-2\langle v_{\eta,k},\hat{v}\rangle = 4-4h_{\eta,k}^{1} \notag \\
%& = 4-4\langle h_{\eta,k},\hat{h}\rangle \notag\\
& = 4-4\sqrt{1-\sum\nolimits_{i=2}^{2d}\big(h_{\eta,k}^{(i)}\big)^2}=  \frac{16\sum\nolimits_{i=2}^{2d}\big(h_{\eta,k}^{(i)}\big)^2}{4+4\sqrt{1-\sum\nolimits_{i=2}^{2d}\big(h_{\eta,k}^{(i)}\big)^2}} \leq 3\sum\nolimits_{i=2}^{2d}\big(h_{\eta,k}^{(i)}\big)^2,
\end{align}
where the last inequality holds since $k$ is large enough such that $\sum\nolimits_{i=2}^{2d}\big(h_{\eta,k}^{(i)}\big)^2$ is sufficiently small. By Propositions~\ref{Time_Saddle},~\ref{Time_Deterministic}, and~\ref{Time-Optimal}, the total iteration number is
\begin{align}\label{T}
N & =  N_1+N_2+N_3.
\end{align}
To explicitily bound $N$ in \eqref{T} in terms of sample size n,  we consider
\begin{align}
N_1&= \frac{\eta^{-1}}{2(\lambda_1-\lambda_2)}\log\left(\frac{2\eta^{-1}\delta^2(\lambda_1-\lambda_2)}{\Phi^{-1}\left(\frac{1+\nu}{2}\right)^2\beta^2_{12}} +1\right),  \label{N_1} \\
N_2 & = \frac{\eta^{-1}}{2(\lambda_1-\lambda_2)}\log\frac{1-\delta^2}{\delta^2},  \label{N_2}\\
N_3 & = \frac{\eta^{-1}}{2(\lambda_1-\lambda_2)}\log\left(\frac{4(\lambda_1-\lambda_2)\delta^2}{(\lambda_1-\lambda_2)\epsilon\eta^{-1}-4d\max\limits_{1\leq i \leq d}\beta_{i1}^2}\right) . \label{N_3}
\end{align}
Given a small enough $\epsilon,$ we choose $\eta$ as follow:
\begin{align}
\eta \asymp \frac{\epsilon (\lambda_1-\lambda_2)}{d\max_{1\leq i \leq d} \beta_{i1}^2}\label{eta}.
\end{align}
Combining the above sample complexities~\eqref{N_1},~\eqref{N_2},~\eqref{N_3}, and~\eqref{eta}, we get
\begin{align}
N=O\left[ \frac{d}{\epsilon (\lambda_1-\lambda_2)^2}\log\left(\frac{d}{\epsilon}\right) \right].
\end{align}
By Proposition~\ref{Time-Optimal} with \eqref{error}, after at most $N$ iterations, we have 
\begin{align*}
\norm{u_{\eta,n}-\hat{u}}_2^2 + \norm{v_{\eta,n}-\hat{v}}_2^2\leq 3\norm{h_{\eta,n}-\hat{h}}_2^2\leq 3\epsilon,
\end{align*}
with probability at least $\frac{3}{4}$.
\end{proof}

\end{document}